%% file: colm2026_conference.tex
\documentclass{article} 
\PassOptionsToPackage{table}{xcolor}
\usepackage[final]{colm2026_conference} 

\usepackage[textsize=tiny]{todonotes}
\usepackage{inconsolata}
\usepackage{multirow}
\usepackage{multicol}
\usepackage{tabularx}
\usepackage{amssymb}
\usepackage{mathtools}
\usepackage{enumitem}
\usepackage{amsthm}
\usepackage{bbm}
\usepackage{comment}
\usepackage{caption}
\usepackage{subcaption}
\usepackage{wrapfig}
\usepackage{pifont}
\usepackage{tabularx}
\usepackage{lettrine}
\usepackage{yfonts} 
\usepackage{booktabs}
\PassOptionsToPackage{table}{xcolor}
\usepackage[table,svgnames,x11names]{xcolor}
\usepackage{colortbl}
\usepackage{float}
\usepackage{fontawesome}
\usepackage{makecell}
\usepackage{adforn}
\usepackage{lineno}
\usepackage{titlesec}

\input{math_commands.tex}

\input{commands}

\usepackage{hyperref}
\usepackage{url}

\title{Language Models that Think, Chat Better} 

\author{Adithya Bhaskar\thanks{Equal contribution.} $^{\,\,\mathcal{P}}$ \qquad\qquad\qquad\qquad\quad Xi Ye$^{*\mathcal{PA}}$ \qquad\qquad\qquad\qquad\quad Danqi Chen$^{\mathcal{P}}$ \\
$^{\mathcal{P}}$ Princeton Language and Intelligence, Princeton University \\
$^{\mathcal{A}}$ University of Alberta
 \\
\texttt{adithyab@cs.princeton.edu}
}

\titlespacing*{\section}
{0pt}{0.65ex}{0.3ex}

\titlespacing*{\subsection}
{0pt}{0.35ex}{0.2ex}

\begin{document}

\ifcolmsubmission
\linenumbers
\fi

\maketitle

\input{sections/abstract}

\input{sections/intro}
\input{sections/training}
\input{sections/experiments}

\input{sections/analysis}
\input{sections/related_work}
\input{sections/conclusions}

\bibliography{colm2026_conference}
\bibliographystyle{colm2026_conference}

\appendix

\input{appendices/hyperparameters}
\input{appendices/reward_hacking}
\input{appendices/more_results}
\input{appendices/prompts}
\input{appendices/star_po}

\input{appendices/faqs}

\end{document}

%% file: math_commands.tex
\usepackage{amsmath,amsfonts,bm}

\def\eqref#1{equation~\ref{#1}}

\def\1{\bm{1}}

\DeclareMathAlphabet{\mathsfit}{\encodingdefault}{\sfdefault}{m}{sl}
\SetMathAlphabet{\mathsfit}{bold}{\encodingdefault}{\sfdefault}{bx}{n}



%% file: commands.tex
\newcommand\methodname{RLMT}

\usepackage[most,skins,theorems]{tcolorbox}

%% file: sections/abstract.tex
\begin{abstract}

Reinforcement learning with verifiable rewards (RLVR) trains language models to use long chain-of-thought reasoning (CoT) in domains like mathematics and code with rule-based verifiers.
However, long CoT learned through RLVR does not generalize well to open-ended tasks—such as writing essay outlines or making meal plans—where humans reason routinely.
This paper establishes the benefits of long CoT for general-purpose chat capabilities and introduces \textbf{RL} with \textbf{M}odel-rewarded \textbf{T}hinking (\textbf{RLMT})\footnote{We release our code and data publicly at \url{https://github.com/princeton-pli/RLMT}.}, which pushes RLVR beyond verifiable domains.
Using diverse real-world prompts, \methodname{} requires LMs to generate long CoT reasoning before responding, and optimizes them with online RL against a preference-based reward model used in RLHF.
Across 40 training runs on Llama-3.1-8B and Qwen-2.5-7B (both base and instruct) and multiple optimization algorithms (DPO, PPO, and GRPO), RLMT consistently outperforms standard RLHF pipelines.
This includes substantial gains of 3--7 points on three chat benchmarks (AlpacaEval2, WildBench, and ArenaHardV2), along with 1--3 point improvements on other tasks like creative writing and general knowledge.
\methodname{} can also be applied directly to base models without an SFT stage, akin to DeepSeek-R1-Zero. 
Remarkably, with only 7K prompts, Llama-3.1-8B base trained with our RLMT recipe outperforms Llama-3.1-8B-Instruct post-trained with a complex multi-staged pipeline with 25M+ examples.
We close with qualitative and quantitative analyses of how trained models plan their responses.
Our results rethink the post-training pipeline and call upon future work to understand and employ thinking more broadly.

\end{abstract}

%% file: sections/intro.tex
\section{Introduction}

Thinking through the consequences of one's actions---and 
revising them when needed—is a defining feature of human intelligence (often called ``system 2 thinking'',~\citet{kahneman2011thinking}).
It has also become a central aspiration for large language models (LLMs).
Recent progress toward this goal has been driven by reasoning models trained through reinforcement learning with verifiable rewards~\cite[RLVR;][]{lambert2024tulu3,deepseekai2025deepseekr1}. 
In RLVR, models are optimized with automatically checkable rewards from domains such as mathematics and code, encouraging them to reason with a long chain-of-thought~\citep[CoT;][]{nye2021show,wei2022chain} before answering.

\begin{wrapfigure}{l}{.4\textwidth}
    \centering
    \vspace*{-1em}
    \includegraphics[width=1.0\linewidth,trim=0 0 0 0,clip]{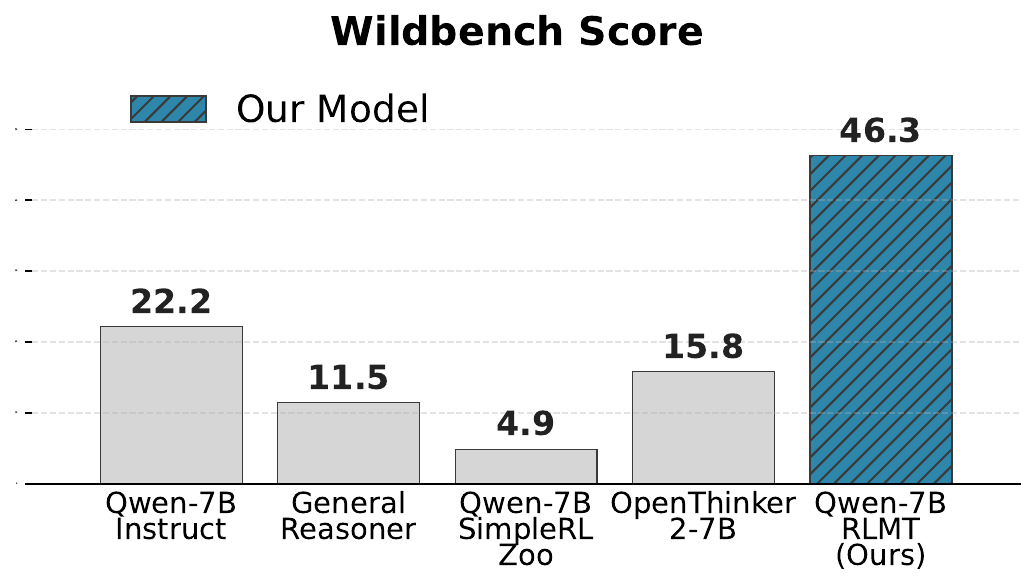}
    \vspace*{-1.5em}
    \caption{Score on WildBench~\citep{lin2025wildbench}. Thinking models trained only on verifiable domains do not generalize well to general-purpose chat.
    }
    \vspace*{-1.2em}
    \label{fig:rlvr_on_chat}
\end{wrapfigure}

RLVR has been applied extensively to math, coding~\citep{deepseekai2025deepseekr1,zeng2025simplerl}, STEM problems~\citep{ma2025generalreasoner}, puzzles and games~\citep{chen2025enigmata,liu2025synlogic,stojanovski2025reasoning,liu2025prorl}, demonstrating strong performance and incentivizing LMs to use sophisticated reasoning patterns such as branching and reflection.
However, we find that skills acquired from RLVR do not generalize to broader tasks requiring open-ended reasoning—such as writing emails, drafting essay outlines, and making to-do lists—where humans also engage in everyday reasoning.
Figure~\ref{fig:rlvr_on_chat} shows that open-source reasoning models trained via math-focused RLVR lag behind standard instruction-tuned models on WildBench~\citep{lin2025wildbench}, a widely used chat benchmark with diverse user queries. 
Complementary studies report limited generalization of RLVR-trained models to reasoning tasks beyond verifiable domains~\citep{huan2025does,zhou2025does}. Such gap raises an important question: whether and how long CoT benefits open-ended reasoning?

\begin{wrapfigure}{r}{.4\linewidth}
    \centering
    \vspace*{-1em}
    \includegraphics[width=1.0\linewidth,trim=0 0 0 0,clip]{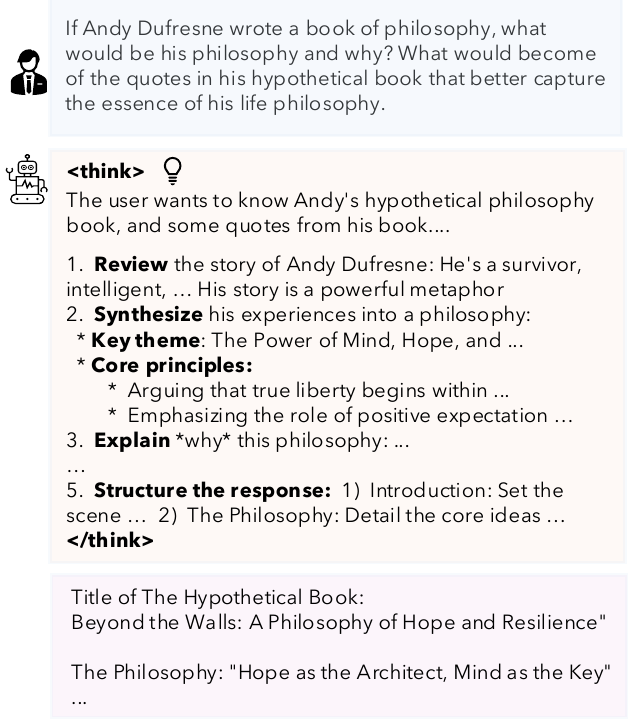}
    \vspace*{-1.5em}
    \caption{Example reasoning trace generated by an LM trained with RLMT for an open-ended query.}
    \vspace*{-1em}
    \label{fig:intro_exs}
\end{wrapfigure}

\begin{figure}[t]
    \centering
    \includegraphics[width=0.9\linewidth]{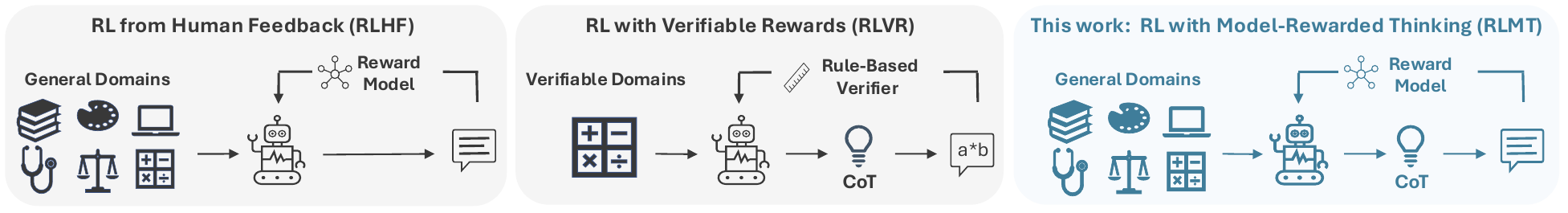}
    \vspace{-0.5em}
    \caption{We train LMs with long chain-of-thought on diverse, general user prompts through reinforcement learning with a reward model. RLMT allows models to think compared to RLHF, and extends RLVR to broader, open-ended tasks. }
    \vspace{-1.5em}
    \label{fig:intro}
\end{figure}

This paper pushes the RLVR paradigm beyond verifiable domains, and introduces \textbf{R}einforcement \textbf{L}earning with \textbf{M}odel-rewarded \textbf{T}hinking (\textbf{RLMT}).
As in Figure~\ref{fig:intro}, \methodname{} trains LMs to generate long CoT reasoning before final answers, using online RL algorithms such as GRPO~\citep{shao2024deepseekmath}. 
Unlike RLHF~\citep{ziegler2020finetuning,ouyang2022training}, this
design encourages explicit reasoning. Compared to RLVR relying on rule-based rewards tied to ground-truth answers, RLMT only requires prompts and uses reward models trained on human preference data over diverse prompts, as in RLHF, to evaluate responses. \methodname{} recipe is surprisingly effective across a wide range of tasks, enabling long CoT for open-ended tasks (see Figure~\ref{fig:intro_exs} for an example).

We systematically study the benefits of long CoT enabled by \methodname{} through extensive experiments covering various optimization algorithms (on-policy DPO, PPO, and GRPO), model families (Llama-3.1-8B and Qwen-2.5-7B), and training configurations.
We apply \methodname{} on both base models and those warm-started with supervised fine-tuning (SFT), akin to R1-zero and R1~\citep{deepseekai2025deepseekr1}.
Across all settings, we find consistent benefits of long CoT: \methodname{} produces sizable improvements over standard RLHF, with average gains of 3--7 points on chat benchmarks and 1--3 points on other tasks including creative writing and general knowledge.
Notably, Llama-3.1-8B-Instruct trained with RLMT (GRPO) scores 58.7 on AlpacaEval2 and 50.4 on WildBench (Table~\ref{tab:grpo-benchmarks}), surpassing models $10\times$ larger that do not use long CoT (Llama-3.1-70B-Instruct), and even outperforming GPT-4o~\citep{openai2024gpt4o} on WildBench.

Remarkably, \methodname{} also delivers substantial gains when applied directly to base models without an SFT stage, revealing insights into the open-ended reasoning capabilities already embedded in base models.
In this setting, \methodname{} achieves average chat scores of 15.6 on Llama-3.1-8B and 29.0 on Qwen-2.5-7B (Table~\ref{tab:grpo-benchmarks})—higher than Llama-3.1-8B-Instruct and Qwen-2.5-7B-Instruct by more than 5 points, despite the latter relying on far more complex post-training pipelines involving millions of examples, rejection sampling, and iterative preference optimization~\citep{grattafiori2024llama3}.

Our analyses surface several interesting findings.
Comparisons between prompt mixtures suggest that diverse, challenging real-world prompts better elicit reasoning skills during RL than simpler synthetic prompts.
We also observe an interesting reversal in model performance: while Llama-3.1 underperforms Qwen-2.5 before RL, the trend reverses after \methodname{} training, suggesting that \methodname{} can reinforce latent capabilities even when not fully optimized during pretraining or SFT.
Lastly, we observe emergent reasoning behaviors in Llama-3.1-8B after RL, such as constraint enumeration and iterative refinement, accompanied by increased CoT length.
Together, our results demonstrate the promise of the long CoT paradigm and open productive directions for future work.

%% file: sections/training.tex
\section{Reinforcement Learning with Model-rewarded Thinking}
\label{sec:training}

\subsection{Background}
We first set up the preliminary background on two LM training paradigms, RL from Human Feedback (RLHF;~\citet{ziegler2020finetuning} and RL with Verifiable Rewards (RLVR;~\citet{lambert2024tulu3}.

\paragraph{RLHF.} The goal of RLHF is to align LM outputs with human preferences. Let $\pi_\theta$ denote a language model with parameters $\theta$.
Given a prompt $x \sim \mathcal{X}$, the LM generates a response $y \sim \pi(\cdot \mid x)$.
Let  $r$ denote a reward function that assigns a scalar score to response $y$ for prompt $x$, i.e., $r(y, x) \in \mathbb{R}$. In practice, $r$ is instantiated as a reward model trained on human preference data, so that higher scores correspond to outputs better aligned with human judgments. RLHF optimizes $\theta$ to maximize the expected reward of responses generated from $\pi_\theta$: 

\footnotesize
\begin{equation}
\max_{\theta} \mathbbm{E}_{x\sim\mathcal{X}}\left[\mathbbm{E}_{y~\sim \pi_{\theta} (\cdot|x)} r(x,y)\right]
\end{equation}
\normalsize

\paragraph{RLVR.} RLVR has become the de facto method for training LMs in domains where ground-truth verification is possible, such as mathematics or code. RLVR modifies the RLHF framework by replacing the model-based reward $r$ with a verification function; for example, the indicator function $\mathbbm{1}\{y = y^*\}$ against a ground-truth answer $y^*$.  
In practice, the verification function may go beyond simple equality checks (e.g., using unit tests for code generation).

Another distinction of RLVR from RLHF is that the LMs usually first produce a reasoning trace $z\sim\pi_{\theta}(\cdot\mid x)$ before a response $y\sim\pi_{\theta}(\cdot\mid x, z)$~\citep{deepseekai2025deepseekr1}, instead of directly generating responses. The optimization objective then maximizes the expected correctness of the final response:

\footnotesize
\begin{equation}
\max_{\theta} \mathbbm{E}_{x \sim \mathcal{X}} \left[ \mathbbm{E}_{(y,z) \sim \pi_\theta(\cdot\mid x)} \mathbbm{1}\{y=y^*\} \right].
\end{equation}
\normalsize

For both RLHF and RLVR, there is a variety of RL or on-policy preference learning algorithms that can be used. In this work, we focus on three widely adopted methods (DPO, PPO, and GRPO). We provide more details in Appendix~\ref{app:brief-dpo-ppo-grpo}.

\subsection{RLMT: Combining RLHF and RLVR}
\label{sec:setup}

While recent RLVR models achieve strong results in formal domains, they exhibit limited generalization to broader reasoning problems~\citep{huan2025does,zhou2025does} and chat benchmarks (see Figure~\ref{fig:rlvr_on_chat} and results in \S\ref{sec:analysis}).  Meanwhile, planning and reasoning do help humans perform a wide range of day-to-day tasks.

We propose \textbf{R}einforcement \textbf{L}earning with \textbf{M}odel-rewarded \textbf{T}hinking (RLMT) to employ broad supervision for open-ended tasks. RLMT uses the following objective: 

\footnotesize
\begin{equation}
\label{eq:objective}
\max_{\theta} \mathbbm{E}_{x\sim\mathcal{X}}\left[\mathbbm{E}_{(y,z)~\sim \pi_{\theta} (\cdot|x)} r(y,x)\right].
\end{equation}
\normalsize

As in Eq (\ref{eq:objective}), RLMT requires LMs to generate a reasoning trace $z$ before producing the final response $y$, which differs from RLHF, and uses a reward model $r$ to score responses, rather than rule-based verification as in RLVR. We study \textbf{several key design choices} for \methodname{}:

\paragraph{Training algorithm.} We experiment with different RL algorithms: on-policy DPO~\citep{rafailov2023direct}, PPO~\citep{schulman2017proximal}, and GRPO~\citep{shao2024deepseekmath}.\footnote{Unlike standard DPO using a static preference dataset, we sample preference pairs using the policy model.}
The choice of training algorithm leads to different performance outcomes (\S\ref{sec:main_results}).
Our best-performing models are trained with GRPO, but our models remain better than baselines in all settings.

\paragraph{Reward model.} 
We adopt \texttt{Skywork-v1-Llama-3.1-8B-v0.2}~\citep{liu2024skywork} as our reward model $r$, which has shown strong performance on reward benchmarks~\citep{liu2024rm} and downstream applications~\citep{malik2025rewardbench2advancingreward}.
We find that having a strong reward model is instrumental for \methodname{} (ablations in Section~\ref{sec:analysis}).

\paragraph{Prompt mixture.}
We construct the prompt distribution from diverse, real-world user requests.
Concretely, we use $7.5$k prompts from the \emph{WildChat-IF} subset of the T\"ulu3 SFT mixture.\footnote{\url{https://huggingface.co/datasets/allenai/tulu-3-wildchat-if-on-policy-8b}}
This subset prioritizes conversational prompts sampled from WildChat~\citep{zhao2024wildchat}, covering a wide range of realistic user queries.
In contrast to the full T\"ulu-3M SFT mixture that contains a high proportion of math and jailbreak prompts, using WildChat-IF allows us to better capture general usage.
Analysis shows that this choice improves general-purpose chat performance over alternatives such as UltraChat~\citep{cui2024ultrafeedback}; see \S\ref{sec:analysis} for details.

\subsection{Warm-start SFT Training and ``Zero'' training}
\label{sec:elicithinking}

Since the LMs we use do not naturally adopt the desired thinking format, we try two methods to elicit this behavior:
(1) warm-starting with supervised fine-tuning (SFT), and
(2) directly prompting base models without SFT (the ``Zero'' approach of \citet{deepseekai2025deepseekr1}).

\paragraph{Warm-start thinking with SFT.}
We begin by teaching models the desired thinking format via supervised fine-tuning (SFT).
Specifically, we sample $6$k prompts from the T\"ulu3 SFT mixture (disjoint from those prompts used for RLMT) for SFT. We generate responses using Gemini 2.5 Flash (0417 Preview), a popular teacher model in recent approaches that distill reasoning behavior from reasoning models~\citep{muennighoff2025s1,guha2025openthoughts}.
Since Gemini's CoT is not accessible, we prompt it to produce a simulated thinking trace before the final response. We additionally experiment with SFT data generated by GPT-4.1-mini and observe similar results and trends (see \S\ref{sec:warmstart_data}). We list details of hyperparameters in Appendix~\ref{app:hyperparameters} and prompt formats in Appendix~\ref{app:prompts}.

\paragraph{Zero training with base models.} We also directly apply \methodname{} to base models without a warm-start, which we refer to as the \emph{Zero} setting. Concretely, we experiment with Llama-3.1-8B~\citep{grattafiori2024llama3} and Qwen-2.5-7B~\citep{qwen2025qwen25}, neither of which has undergone post-training.
In this case, we elicit the desired output structure by prepending a fixed instruction prefix (``\texttt{A conversation between User and Assistant...}''; see Appendix~\ref{app:prompts}).
The subsequent RL training procedure is otherwise identical to that for \methodname{}.


%% file: sections/experiments.tex
\section{Thinking Benefits Open-ended Reasoning}
\label{sec:main_results}

\subsection{Settings, Benchmarks and Evaluated Models}
\paragraph{Setting and models.} We evaluate RLMT in two settings: (1) applied to models after SFT warm-start (Tables~\ref{tab:grpo-benchmarks} and \ref{tab:dpo-ppo-benchmarks}), and (2) applied directly to base models (the ``zero'' setting; Tables~\ref{tab:grpo-benchmarks} and \ref{tab:dpo-ppo-benchmarks}).
We apply setting (1) with SFT on top of the Base and Instruct versions of Llama-3.1-8B and Qwen2.5-7B. Following~\citet{deepseekai2025deepseekr1}, we apply the zero training (2) only to base models.
In Table~\ref{tab:grpo-benchmarks}, we report GRPO as our main results, since it achieves the best overall performance and serves as the basis for our analysis. We provide results with DPO and PPO in Table~\ref{tab:dpo-ppo-benchmarks} for comparison.

\paragraph{Benchmarks.}
We evaluate our models on a suite of 7 benchmarks spanning general chat, creative writing, instruction following, and general knowledge---these are chosen to represent a meaningful selection of broadly applicable tasks.
We list the benchmarks below:

\begin{wraptable}{r}{0.425\textwidth}
    \renewcommand{\tabcolsep}{1.0mm}
    \fontsize{7.75}{7.75}\selectfont
    \centering
    \vspace{0em}
    \caption{Comparison of Llama-3.1-8B-Instruct-RLMT with strong open-source and closed models, including GPT-4o and Claude-3.7-Sonnet (a thinking model).}
    \vspace{-1em}
    \label{tab:vs-frontier}
    \begingroup
    \begin{tabular}{lrrrrr}
        \toprule
        \textbf{\emph{Model}} & \textbf{\emph{Avg.}} & \textbf{\emph{WB}} & \textbf{\emph{AE2}} & \textbf{\emph{AH2}} & \textbf{\emph{CWv3}} \\
        \midrule
        \multicolumn{5}{l}{\textit{\textbf{Our model}}} \\
        \rowcolor{green!14} L3.1-8B-I-RLMT & \underline{54.1} & \textbf{50.4} & \textbf{58.7} & 22.9 & \underline{84.3}\\
        \multicolumn{5}{l}{\textit{\textbf{Other models}}} \\
        L3.1-70B-Instruct & 32.1 & 16.3 & 42.0 & 10.6 & 59.4\\
        Q2.5-72B-Instruct & 45.2 & 44.4 & 50.2 & 19.9 & 66.3\\
        GPT-4o & 53.2 & 46.2 & 56.5 & \underline{32.1} & 77.8\\
        \makecell{Claude3.7-Sonnet} & \textbf{58.9} & \underline{47.8} & \underline{58.1} & \textbf{39.3} & \textbf{90.3}\\
        \bottomrule
    \end{tabular}
    \vspace{-1em}
    \endgroup
\end{wraptable}

\vspace{-0.5em}
\begin{enumerate}[noitemsep,leftmargin=15px]
    \item \textbf{Chat.} We include the widely used  \textbf{WildBench (WB)}~\citep{lin2025wildbench}, \textbf{AlpacaEval 2 (AE2)}~\citep{dubois2024lengthcontrolled}, and \textbf{ArenaHardV2 (AH2)}~\citep{li2024from,li2024fromb} for chat evaluation. AE2 and AH2 use a free-form judgment procedure, whereas the WB relies on carefully crafted rubrics. 
    \item \textbf{Creative writing.} We augment these benchmarks with \textbf{CreativeWritingV3 (CWv3)}~\citep{samuel2025creativewritingv3} to evaluate the creative writing abilities of our models. The WB score ranges from -100 to 100, while the other three range from 0 to 100.
    \item \textbf{Instruction following.} We use the recently published \textbf{IFBench (IF\textsubscript{Ben})} benchmark~\citep{pyatkin2025generalizing} to produce a score ranging from 0--100.
    \item \textbf{General knowledge.} We evaluate our models on \textbf{MMLU-Redux (MMLU\textsubscript{R})}~\citep{gema2025are} and \textbf{PopQA}~\citep{mallen2022when} to test general and long-tail knowledge, respectively. The resulting scores span 0--100. 
\end{enumerate}
We provide more details on the evaluation process (e.g., judge model, length control) along with four more benchmarks, including math and logical puzzles, in Appendix~\ref{sec:more_results}.

\paragraph{Baselines and algorithms.}
We pair every RLMT model with an RLHF baseline under the same training setup, differing only in the absence of thinking.
To rigorously isolate the effect of integrating long CoT in post-training, we construct a matched set of \emph{non-thinking} baselines trained with RLHF paradigm.
For every thinking model in any setting, we train a corresponding non-thinking model that follows the same setting without thinking. Concretely, we still take prompt-response pairs distilled from Gemini thinking for a fair comparison, but we removed the thinking trace in this case.
We evaluate our models and baselines with DPO, PPO, and GRPO (more details in Appendix~\ref{app:hyperparameters}).

\begin{table}[t]
\centering
\small
\renewcommand{\tabcolsep}{1.0mm}
\fontsize{7.5}{7.5}\selectfont
\vspace{-0.75em}
\caption{GRPO results for models trained from Llama-3.1-8B and Qwen2.5-7B (base and instruct) in both warm-started and zero settings. \faLightbulbO\ shows whether thinking was enabled, with $\checkmark$ denoting RLMT models and $\times$ denoting RLHF models. The best numbers are \textbf{bolded} in each category.  Our main focus is on chat benchmarks: WildBench (WB), AlpacaEval2 (AE2), and ArenaHardV2 (AH2). When evaluating untrained base models, we prompt them with both thinking and nonthinking templates (tpl). Thinking models outperform non-thinking baselines, especially on chat and creative writing.}
\vspace{-0.75em}
    \label{tab:grpo-benchmarks}
    \begingroup
\begin{tabular}{lllrrrrrrrrrrr}
\toprule
\textbf{\emph{Backbone}} & \textbf{\emph{Training}} & \textbf{\emph{\faLightbulbO }} & \textbf{\emph{WB}} & \textbf{\emph{AE2}} & \textbf{\emph{AH2}} & \textbf{\emph{Avg\textsubscript{Chat}}} & \textbf{\emph{}} & \textbf{\emph{CWv3}} & \textbf{\emph{PopQA}} & \textbf{\emph{IF\textsubscript{Ben}}} & \textbf{\emph{MMLU\textsubscript{R}}} & \textbf{\emph{}} & \textbf{\emph{Avg}} \\
\cmidrule{1-7} \cmidrule{9-12} \cmidrule{14-14}
\multicolumn{14}{l}{\underline{\textit{SFT Warm-Started Models}}} \\
Llama-3.1-8B & + SFT & $\times$ & \cellcolor{blue!14}-10.6 & \cellcolor{blue!14}26.8 & \cellcolor{blue!14}5.1 & \cellcolor{blue!14}7.1 &  & 75.2 & 26.4 & 15.6 & 59.7 &  & 28.3 \\
\rowcolor{gray!14}
 &  & $\checkmark$ & \cellcolor{blue!14}-1.6 & \cellcolor{blue!14}29.7 & \cellcolor{blue!14}6.5 & \cellcolor{blue!14}11.5 &  & 75.1 & \textbf{30.5} & 17.0 & 61.1 &  & 31.2 \\
 &   + GRPO & $\times$ & \cellcolor{blue!14}33.2 & \cellcolor{blue!14}46.7 & \cellcolor{blue!14}\textbf{16.3} & \cellcolor{blue!14}32.1 &  & \textbf{84.2} & 24.5 & \textbf{17.7} & 61.2 &  & 40.5 \\
\rowcolor{green!14}
\multicolumn{2}{l}{\textbf{\textit{Llama-3.1-8B-RLMT}}} & $\checkmark$ & \cellcolor{blue!28}\textbf{38.1} & \cellcolor{blue!28}\textbf{52.3} & \cellcolor{blue!28}15.9 & \cellcolor{blue!28}\textbf{35.4} &  & 80.9 & 30.3 & 15.6 & \textbf{61.7} &  & \textbf{42.1} \\
\cmidrule{1-7} \cmidrule{9-12} \cmidrule{14-14}
Qwen2.5-7B & + SFT & $\times$ & \cellcolor{blue!14}-0.9 & \cellcolor{blue!14}28.8 & \cellcolor{blue!14}8.0 & \cellcolor{blue!14}12.0 &  & 61.9 & 21.1 & 19.0 & 65.2 &  & 29.0 \\
\rowcolor{gray!14}
 &  & $\checkmark$ & \cellcolor{blue!14}12.0 & \cellcolor{blue!14}33.9 & \cellcolor{blue!14}10.6 & \cellcolor{blue!14}18.8 &  & \textbf{69.5} & 22.0 & 21.1 & 62.1 &  & 33.0 \\
 &  + GRPO & $\times$ & \cellcolor{blue!14}28.9 & \cellcolor{blue!14}51.0 & \cellcolor{blue!14}13.1 & \cellcolor{blue!14}31.0 &  & 60.9 & \textbf{24.4} & 19.7 & \textbf{72.8} &  & 38.7 \\
\rowcolor{green!14}
\multicolumn{2}{l}{\textbf{\textit{Qwen2.5-7B-RLMT}}} & $\checkmark$ & \cellcolor{blue!28}\textbf{31.0} & \cellcolor{blue!28}\textbf{54.0} & \cellcolor{blue!28}\textbf{19.1} & \cellcolor{blue!28}\textbf{34.7} &  & 65.7 & 22.8 & \textbf{21.8} & 67.3 &  & \textbf{40.2} \\
\cmidrule{1-7} \cmidrule{9-12} \cmidrule{14-14}
Llama-3.1-8B-Instruct &  & $\times$ & \cellcolor{blue!14}-7.0 & \cellcolor{blue!14}32.1 & \cellcolor{blue!14}5.1 & \cellcolor{blue!14}10.1 &  & 55.0 & \textbf{36.4} & 23.8 & 70.0 &  & 30.8 \\
 & + SFT & $\times$ & \cellcolor{blue!14}12.1 & \cellcolor{blue!14}33.5 & \cellcolor{blue!14}9.9 & \cellcolor{blue!14}18.5 &  & 78.5 & 31.1 & 23.5 & 64.8 &  & 36.2 \\
\rowcolor{gray!14}
 &  & $\checkmark$ & \cellcolor{blue!14}14.3 & \cellcolor{blue!14}34.5 & \cellcolor{blue!14}9.5 & \cellcolor{blue!14}19.4 &  & 78.7 & 32.5 & \textbf{24.8} & 70.6 &  & 37.8 \\
 &   + GRPO & $\times$ & \cellcolor{blue!14}42.0 & \cellcolor{blue!14}45.6 & \cellcolor{blue!14}19.9 & \cellcolor{blue!14}35.8 &  & 83.6 & 31.8 & 21.1 & \textbf{71.7} &  & 45.1 \\
\rowcolor{green!14}
\multicolumn{2}{l}{\textbf{\textit{Llama-3.1-8B-Instruct-RLMT}}} & $\checkmark$ & \cellcolor{blue!28}\textbf{50.4} & \cellcolor{blue!28}\textbf{58.7} & \cellcolor{blue!28}\textbf{22.9} & \cellcolor{blue!28}\textbf{44.0} &  & \textbf{84.3} & 34.0 & 22.1 & 70.0 &  & \textbf{48.9} \\ 
\cmidrule{1-7} \cmidrule{9-12} \cmidrule{14-14}
Qwen2.5-7B-Instruct &  & $\times$ & \cellcolor{blue!14}22.2 & \cellcolor{blue!14}37.1 & \cellcolor{blue!14}10.0 & \cellcolor{blue!14}23.1 &  & 49.8 & 22.2 & \textbf{28.2} & \textbf{75.4} &  & 35.0 \\
 & + SFT & $\times$ & \cellcolor{blue!14}12.6 & \cellcolor{blue!14}28.6 & \cellcolor{blue!14}10.0 & \cellcolor{blue!14}17.1 &  & 65.4 & 21.1 & 19.0 & 67.2 &  & 32.0 \\
\rowcolor{gray!14}
 &  & $\checkmark$ & \cellcolor{blue!14}18.7 & \cellcolor{blue!14}33.7 & \cellcolor{blue!14}10.9 & \cellcolor{blue!14}21.1 &  & 71.0 & 21.8 & 21.8 & 60.5 &  & 34.1 \\
 & + GRPO & $\times$ & \cellcolor{blue!14}37.4 & \cellcolor{blue!14}41.6 & \cellcolor{blue!14}16.3 & \cellcolor{blue!14}31.8 &  & 72.6 & 21.9 & 17.0 & 73.1 &  & 40.0 \\
\rowcolor{green!14}
\multicolumn{2}{l}{\textbf{\textit{Qwen2.5-7B-Instruct-RLMT}}} & $\checkmark$ & \cellcolor{blue!28}\textbf{46.3} & \cellcolor{blue!28}\textbf{50.5} & \cellcolor{blue!28}\textbf{20.8} & \cellcolor{blue!28}\textbf{39.2} &  & \textbf{75.6} & \textbf{22.5} & 20.1 & 71.5 &  & \textbf{43.9} \\
\midrule
\multicolumn{14}{l}{\underline{\textit{Zero Training (No SFT)}}} \\
Llama-3.1-8B & Base (nonthink tpl) & $\times$ & \cellcolor{blue!14}-87.6 & \cellcolor{blue!14}1.7 & \cellcolor{blue!14}0.8 & \cellcolor{blue!14}-28.4 &  & 30.2 & 25.2 & 17.0 & 36.6 &  & 3.4 \\
\rowcolor{gray!14}
 & Base (think tpl) & $\checkmark$ & \cellcolor{blue!14}-88.2 & \cellcolor{blue!14}1.6 & \cellcolor{blue!14}0.6 & \cellcolor{blue!14}-28.7 &  & 31.8 & 20.7 & 16.7 & 26.5 &  & 1.4 \\
 &  + GRPO   & $\times$ & \cellcolor{blue!14}-4.8 & \cellcolor{blue!14}29.8 & \cellcolor{blue!14}4.5 & \cellcolor{blue!14}9.8 &  & 47.5 & 29.5 & 16.0 & 55.2 &  & 25.4 \\
\rowcolor{green!14}
\multicolumn{2}{l}{\textbf{\textit{Llama-3.1-8B-RLMT-Zero}}} & $\checkmark$ & \cellcolor{blue!28}\textbf{7.2} & \cellcolor{blue!28}\textbf{34.0} & \cellcolor{blue!28}\textbf{5.6} & \cellcolor{blue!28}\textbf{15.6} &  & \textbf{49.0} & \textbf{31.2} & \textbf{18.0} & \textbf{56.2} &  & \textbf{28.7} \\
\cmidrule{1-7} \cmidrule{9-12} \cmidrule{14-14}
Qwen2.5-7B & Base (nonthink tpl) & $\times$ & \cellcolor{blue!14}-65.8 & \cellcolor{blue!14}4.5 & \cellcolor{blue!14}2.0 & \cellcolor{blue!14}-19.8 &  & 39.1 & 23.1 & 17.3 & 63.2 &  & 11.9 \\
\rowcolor{gray!14}
 & Base (think tpl) & $\checkmark$ & \cellcolor{blue!14}-68.4 & \cellcolor{blue!14}4.4 & \cellcolor{blue!14}1.5 & \cellcolor{blue!14}-20.8 &  & 36.7 & 23.0 & 17.3 & 59.9 &  & 10.6 \\
 & + GRPO   & $\times$ & \cellcolor{blue!14}13.4 & \cellcolor{blue!14}48.4 & \cellcolor{blue!14}8.8 & \cellcolor{blue!14}23.5 &  & 50.7 & \textbf{25.2} & 16.0 & \textbf{71.9} &  & 33.5 \\
\rowcolor{green!14}
\multicolumn{2}{l}{\textbf{\textit{Qwen2.5-7B-RLMT-Zero}}} & $\checkmark$ & \cellcolor{blue!28}\textbf{22.2} & \cellcolor{blue!28}\textbf{54.0} & \cellcolor{blue!28}\textbf{10.8} & \cellcolor{blue!28}\textbf{29.0} &  & \textbf{54.0} & 24.2 & \textbf{18.0} & 71.8 &  & \textbf{36.4} \\
\bottomrule
\end{tabular}
\vspace{-1.5em}
\endgroup
\end{table}

\subsection{Results}

\paragraph{Thinking models excel in chat and creative writing.}  Table~\ref{tab:grpo-benchmarks} contains results after SFT, and after GRPO for both thinking and non-thinking models.
Thinking models trained with RLMT consistently outperform non-thinking counterparts by 1.5--4 points on average across all benchmarks (Tables~\ref{tab:grpo-benchmarks} 
and \ref{tab:dpo-ppo-benchmarks}).
The gap over baselines is maximum on chat (WildBench and AlpacaEval2): 3--8 points on average.
They are usually also better at creative writing and factuality (PopQA).
To provide a further reference, Table~\ref{tab:vs-frontier} compares our best model, Llama-3.1-8B-Instruct-RLMT, to four strong models.
Remarkably, despite being $10\times$ smaller than the two open source models, our RLMT model outperforms them both by large margins (9--22 points).
It also \textbf{outperforms GPT-4o} on chat and creative writing.
We also compare it to the frontier thinking model Claude-3.7-Sonnet.
(February 2025, rumored to be 150B+ scale and post-trained on millions of examples).
Llama-3.1-8B-Instruct-RLMT \textbf{rivals Claude-3.7-Sonnet (thinking) on AlpacaEval2 and Wildbench}, though it is worse on ArenaHardv2, likely due to the high proportion of math and coding.

\paragraph{RLMT remains effective with PPO and DPO.} Comparing Tables~\ref{tab:grpo-benchmarks} and~\ref{tab:dpo-ppo-benchmarks} in Appendix~\ref{sub:ppo} shows that RLMT performs best with GRPO, but remains effective with other RL objectives ---PPO and DPO---as well.

\paragraph{RLMT can elicit chat capabilities even without SFT.}
The zero training sections of Tables~\ref{tab:grpo-benchmarks} (and \ref{tab:dpo-ppo-benchmarks} in the Appendix) summarize the results in the setting that directly applies RLMT to base models.
With GRPO (Table~\ref{tab:grpo-benchmarks}), the models show large improvements on all benchmarks.
RLMT delivers a further 3-point improvement on average, allowing Qwen-RLMT-Zero to outperform Qwen2.5-7B-Instruct.
Llama-RLMT-Zero performs just under Llama-3.1-8B-Instruct, but outperforms it on chat benchmarks by 5.5 points.



\paragraph{Shifts in relative performance across model families.} An interesting observation is that while Llama-3.1 initially lags behind Qwen-2.5 before going through RL, it surpasses Qwen-2.5 after applying \methodname{}. We hypothesize that this arises because Qwen-2.5 undergoes more extensive tuning targeted at these benchmarks, whereas \methodname{} can reinforce and unlock these capabilities in models like Llama-3.1 that are less optimized initially.


%% file: sections/analysis.tex
\section{Analysis}
\label{sec:analysis}


In this section, we undertake several analyses and ablation studies to gain a better understanding of what affects the performance of the thinking models, and in what aspects precisely they improve.

\subsection{Comparison of our RLMT models against models trained on math}
\label{sub:math-vs-ours}
\begin{table}[t]
    \small
    \centering
    \caption{Comparison of RLMT models with math-trained (thinking) models.}
    \label{tab:math-vs-ours}
    \begingroup
    \begin{tabular}{lrrrrr}
        \toprule
        \textbf{\emph{Model}} & \textbf{\emph{Avg.}} & \textbf{\emph{WB}} & \textbf{\emph{AE2}} & \textbf{\emph{AH2}} & \textbf{\emph{CWv3}} \\
        \midrule
        \multicolumn{6}{c}{\textit{(Models sharing similar backbones)}} \\
        \multicolumn{5}{l}{\textit{\textbf{Our models}}} \\
        Llama-3.1-8B-RLMT & 46.8 & 38.1 & 52.3 & 15.9 & 80.9\\
        Qwen2.5-7B-RLMT & 42.4 & 31.0 & 54.0 & 19.1 & 65.7\\
        \rowcolor{green!14} Llama3.1-8B-Inst-RLMT & \textbf{54.1} & \textbf{50.4} & \textbf{58.7} & 22.9 & \textbf{84.3}\\
        Qwen2.5-7B-Inst-RLMT& 48.3 & 46.3 & 50.5 & 20.8 & 75.6\\
        \multicolumn{5}{l}{\textit{\textbf{Thinking models trained on math}}} \\
        DS-R1-Distill-L-8B & 19.9 & -10.2 & 23.8 & 6.1 & 60.0\\
        Q2.5-7B-SimpleRL-Zoo & 25.5 & 4.9 & 35.4 & 8.5 & 53.2\\
        DS-R1-Distill-Q-7B & 8.2 & -29.8 & 16.2 & 6.3 & 40.1\\
        OpenThinker2-7B & 37.4 & 15.8 & 47.7 & 17.5 & 68.4\\
        $\hookrightarrow$ + RLMT & 40.5 & 21.7 & 52.7 & 18.7 & 68.8\\
        \bottomrule
    \end{tabular}
    \endgroup
\end{table}

Several models have tried replicating DeepSeek-R1's success on math and other reasoning domains via distillation, RL, or a combination of the two. 
Do they perform as well on chat and creative writing?
We compare the four RLMT models we train with the prominent models in this space: (1) DeepSeek-R1-Distill (Llama and Qwen)~\citep{deepseekai2025deepseekr1}, (2) Q2.5-7B-SimpleRL-Zoo (Qwen)~\citep{zeng2025simplerl}, (3) DS-R1-Distill-Q-7B (Qwen)~\citep{deepseekai2025deepseekr1}, and (4) OpenThinker2-7B (Qwen)~\citep{guha2025openthoughts}.
We list these models in Table~\ref{tab:math-vs-ours}.
We see that these models do not perform well on chat or creative writing.
All RLMT models outperform all math models by margins of 10--25 points on average.
While training OpenThinker2-7B---the leading math model---with RLMT yields some improvement, our models remain comfortably better.

\begin{table}[ht]
    \centering
    \small
\renewcommand{\tabcolsep}{1.0mm}
\fontsize{7.75}{7.75}\selectfont
    \caption{Ablation of the GRPO prompt mixture, SFT data source, and reward model. Rows marked with ``$\hookrightarrow$'' denote ablations.  We highlight the main model in light green.}
    \label{tab:rl-data-ablation}
    \begingroup
    \begin{tabular}{lllrrrrrrrrrrr}
\toprule
\textbf{\emph{Ablation}} & \textbf{\emph{Objective}} & \textbf{\emph{\faLightbulbO }} & \textbf{\emph{WB}} & \textbf{\emph{AE2}} & \textbf{\emph{AH2}} & \textbf{\emph{Avg\textsubscript{C}}} & \textbf{\emph{}} & \textbf{\emph{CWv3}} & \textbf{\emph{PopQA}} & \textbf{\emph{IF\textsubscript{Ben}}} & \textbf{\emph{MMLU\textsubscript{R}}} & \textbf{\emph{}} & \textbf{\emph{Avg}} \\
\cmidrule{1-7} \cmidrule{9-12} \cmidrule{14-14}

\multicolumn{12}{l}{\textit{Ablation: using different RL prompt mixture (GRPO prompts)}} \\
\rowcolor{green!14} 
Prompts: WildchatIF & \;   + GRPO   & $\checkmark$ & \cellcolor{blue!28}\textbf{38.1} & \cellcolor{blue!28}\textbf{52.3} & \cellcolor{blue!28}\textbf{15.9} & \cellcolor{blue!28}\textbf{35.4} &  & \textbf{80.9} & 30.3 & 15.6 & \textbf{61.7} &  & \textbf{42.1} \\
\rowcolor{gray!14}
$\hookrightarrow$\ Prompts: UltraFeedback & \;     + GRPO   & $\checkmark$ & \cellcolor{blue!14}{35.5} & \cellcolor{blue!14}{47.3} & \cellcolor{blue!14}{13.8} & \cellcolor{blue!14}{32.2} &  & {78.0} & \textbf{33.3} & {16.0} & 50.7 &  & {39.2} \\
\rowcolor{gray!14}
$\hookrightarrow$\ Prompts: T\"ulu3 Random & \;     + GRPO   & $\checkmark$ & \cellcolor{blue!14}22.2 & \cellcolor{blue!14}42.7 & \cellcolor{blue!14}11.5 & \cellcolor{blue!14}25.5 &  & 76.4 & {31.0} & \textbf{16.3} & {59.5} &  & 37.1 \\

\cmidrule{1-7} \cmidrule{9-12} \cmidrule{14-14}
\multicolumn{12}{l}{\textit{Ablation: SFT warm-start with different data source} (\textbf{GPT-4.1-mini})} \\

 $\hookrightarrow$\ Warm-start: GPT-4.1-mini & + SFT & $\times$ & \cellcolor{blue!14}-5.9 & \cellcolor{blue!14}25.1 & \cellcolor{blue!14}5.3 & \cellcolor{blue!14}8.2 &  & 67.7 & {31.3} & {18.0} & 59.2 &  & 28.7 \\
\rowcolor{gray!14}
 $\hookrightarrow$\ Warm-start: GPT-4.1-mini &  & $\checkmark$ & \cellcolor{blue!14}7.0 & \cellcolor{blue!14}26.6 & \cellcolor{blue!14}7.3 & \cellcolor{blue!14}13.6 &  & 72.9 & \textbf{33.1} & \textbf{19.4} & \textbf{68.2} &  & 33.5 \\
 $\hookrightarrow$\ Warm-start: GPT-4.1-mini & \; + GRPO   & $\times$ & \cellcolor{blue!14}{38.6} & \cellcolor{blue!14}{44.8} & \cellcolor{blue!14}{16.2} & \cellcolor{blue!14}{33.2} &  & {82.1} & 28.2 & 16.3 & {60.5} &  & {41.0} \\
\rowcolor{gray!14}
  $\hookrightarrow$\ Warm-start: GPT-4.1-mini&  & $\checkmark$ & \cellcolor{blue!14}\textbf{40.6} & \cellcolor{blue!14}\textbf{51.9} & \cellcolor{blue!14}\textbf{17.5} & \cellcolor{blue!14}\textbf{36.7} &  & \textbf{82.8} & 31.0 & 16.7 & 56.3 &  & \textbf{42.4} \\
\cmidrule{1-7} \cmidrule{9-12} \cmidrule{14-14}
\multicolumn{12}{l}{\textit{Ablation: Reward model} (\textbf{SkyworkV2 \& ArmoRM})} \\

 $\hookrightarrow$\ RM: SkyworkV2 & \; + GRPO   & $\times$ & \cellcolor{blue!14}{32.6} & \cellcolor{blue!14}{47.8} & \cellcolor{blue!14}{18.8} & \cellcolor{blue!14}{33.1} &  & \textbf{82.2} & 32.6 & 19.4 & 55.6 &  & 41.0 \\
\rowcolor{gray!14}
  $\hookrightarrow$\ RM: SkyworkV2 &  & $\checkmark$ & \cellcolor{blue!14}{\textbf{40.9}} & \cellcolor{blue!14}{\textbf{53.2}} & \cellcolor{blue!14}{\textbf{22.8}} & \cellcolor{blue!14}{\textbf{39.0}} &  & {80.4} & \textbf{33.7} & \textbf{19.7} & 47.7 &  & \textbf{42.6} \\
 $\hookrightarrow$\ RM: ArmoRM & \; + GRPO   & $\times$ & \cellcolor{blue!14}{-10.2} & \cellcolor{blue!14}{30.9} & \cellcolor{blue!14}{5.8} & \cellcolor{blue!14}{8.8} &  & {71.2} & 35.4 & 15.0 & \textbf{58.9} &  & 29.6 \\
\rowcolor{gray!14}
  $\hookrightarrow$\ RM: ArmoRM &  & $\checkmark$ & \cellcolor{blue!14}-5.9 & \cellcolor{blue!14}43.3 & \cellcolor{blue!14}6.9 & \cellcolor{blue!14}14.8 &  & 63.3 & 32.6 & 16.7 & 54.6 &  & 30.2 \\
\bottomrule
\end{tabular}
    \endgroup
\end{table}

\subsection{Ablations}
\paragraph{Impact of prompt mixture.}


We vary the prompt mixture used for GRPO to study its impact on downstream performance.
To avoid confounders due to instruction tuning or due to oddities of the Qwen models (see~\citet{shao2025spurious}), we perform this study on the Llama-3.1-8B.
We experiment with two other data sources in Table~\ref{tab:rl-data-ablation}: UltraFeedback~\citep{cui2024ultrafeedback}, popular for preference optimization, and a randomly sampled subset of the Tulu3 SFT mixture~\citep{lambert2024tulu3}.

We see that the Wildchat-IF subset of T\"ulu-3 outperforms both of these sources: we attribute this to the relatively simple prompts in UltraFeedback and the abundance of math and jailbreak (non-chat) prompts in the unfiltered Tulu SFT mixture. In effect, the prompts used for RL matter: more ``chatty'' and harder prompts lead to more improvements.

\paragraph{Impact of warm-start data.}
\label{sec:warmstart_data}
Since Gemini 2.5 Flash is a ``thinking'' model, one might wonder if our takeaways are an artifact of the specific choice of this model.
To this end, we repeat a subset of our experiments (Llama-3.1-8B $\rightarrow$ SFT $\rightarrow$ GRPO) using GPT-4.1-mini to generate the warm-start data, leaving the prompts and other hyperparameters unchanged.
As Table~\ref{tab:rl-data-ablation} shows, thinking models still outperform non-thinking models, especially on chat benchmarks.
The final numbers achieved by the GRPO thinking models are roughly the same as the ones warm-started on Gemini 2.5 Flash.

\paragraph{Impact of reward model.}
To assess the importance of the reward model, we run RLMT (GRPO) with (1) Skywork-V2~\citep{liu2025skyworkrewardv2}, a newer version of the Skywork reward model with carefully curated training data\footnote{Most of our experiments were conducted before the release of Skywork-V2. We believe some of our results may also be pushed further with this version.} (2) ArmoRM~\citep{wang2024interpretable}, another popular reward model used in alignment research~\citep{meng2024simpo}.
Both are based on Llama-3.1-8B backbones, with Skywork-V2 generally representing a stronger reward model than our default Skywork-V1, and ArmoRM representing a weaker one. 
Table~\ref{tab:rl-data-ablation} summarizes the results with and without thinking. We find: 1) \textbf{Stronger reward models lead to better performance.} The gap between the SkyworkV2 and ArmoRM results is large. A weaker model leads to drops on the non-chat benchmarks---especially with thinking. On the other hand, a strong reward model can maintain performance on non-chat benchmarks while also boosting chat performance.
2) \textbf{RLMT outperforms standard RLHF on chat across RMs.} With both RMs, thinking models continue to outperform their non-thinking counterparts by a large gap on chat benchmarks.


\subsection{How does RL training change model behavior?}

\begin{figure}[t]
\vspace{1em}
    \centering
    \begin{subfigure}[t]{0.48\textwidth}
        \centering
        \includegraphics[width=\linewidth]{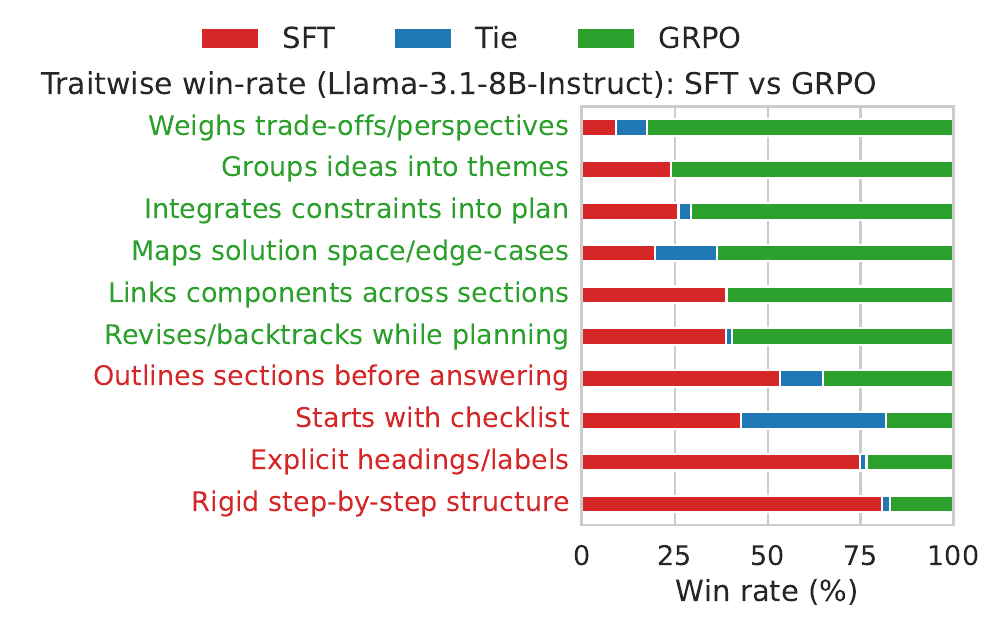}
    \end{subfigure}
    \hfill
    \begin{subfigure}[t]{0.48\textwidth}
        \centering
        \includegraphics[width=\linewidth]{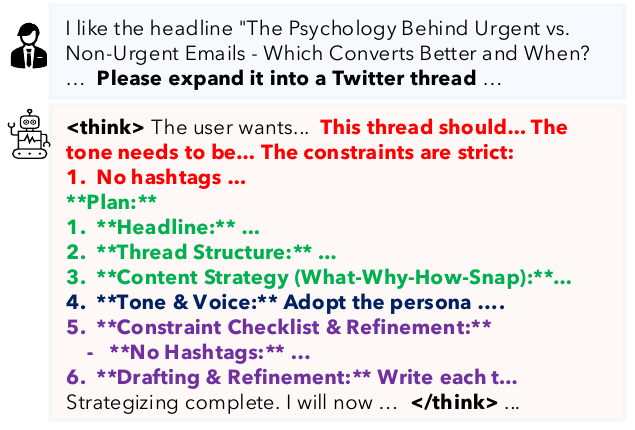}
    \end{subfigure}
    \caption{\textbf{Left}: Traitwise head-to-head win rates for the SFT and GRPO models. Red colors indicate that the trait diminished after GRPO, while green indicates that it increased. \textbf{Right}: Example reasoning behavior. When asked to write a tweet thread, the model first \textcolor{red}{\textbf{maps out the requested constraints }\textcolor{black}{ and then }\textcolor{DarkGreen}{\textbf{plans the tweet progression}}}.
    \textcolor{black}{It also }\textcolor{Purple}{\textbf{runs everything through the checklist and notes necessary refinements }}before producing the final output.
    }
    \label{fig:qual_analysis}
\end{figure}

\paragraph{Qualitative Analysis} We analyze why precisely do the thinking models perform better on chat benchmarks.
To this end, we take the best model from our suite, Llama-3.1-8B-Instruct-RLMT, and the warm-started model before RLMT. We employ the following pipeline to automatically extract the traits that maximally changed between the two versions: (1) We pass all the prompts from WildBench, and the associated thoughts generated by the two models to GPT-4.1-mini, and ask it to extract the traits most prevalent in each thought. (2) We then pass trait-sets from both models in batches of 20 prompts to GPT-4.1-mini to identify the ones consistently more prevalent in one model than the other. (3) We further summarize the identified differences across 10 random batches of 20 prompts. (4) At this point, we have a list of traits that were potentially amplified or suppressed after RLMT. For each, we compute a head-to-head win rate of which model shows the trait more across all 1024 examples of WildBench.



We plot the results in Figure~\ref{fig:qual_analysis} (left). We observe that the SFT model often starts by hierarchically planning out sections, subsections, and then use a checklist to guide the plan. 
On the other hand, Llama-3.1-8B-Instruct-RLMT lists out the relevant constraints and subtopics first, then groups ideas into common themes, only then does it plan out specific details.
We show one (compressed) example of the thinking process in Figure~\ref{fig:qual_analysis} (right) that highlights the identified traits in the model's thoughts.
We also observe that while the SFT model's planning is often linear, the RLMT model's planning is often iterative: it returns to and refines older parts of the plan, e.g., to cross-reference points mentioned elsewhere.
We believe that the strategies reflected in these differences are often the traits exhibited by good writers; it is encouraging that they emerge naturally from the training process.

\begin{figure}[ht]
    \centering
    \begin{subfigure}[t]{0.48\textwidth}
        \centering
        \includegraphics[width=\linewidth]{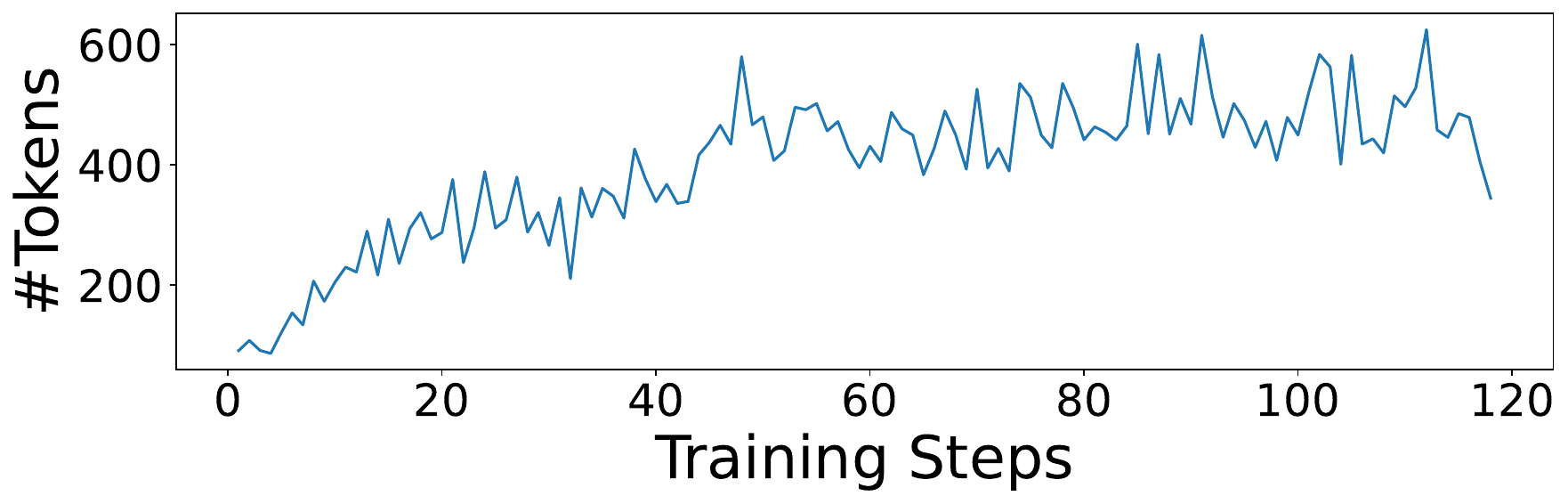}
        \caption{Thoughts}
    \end{subfigure}
    \hfill
    \begin{subfigure}[t]{0.48\textwidth}
        \centering
        \includegraphics[width=\linewidth]{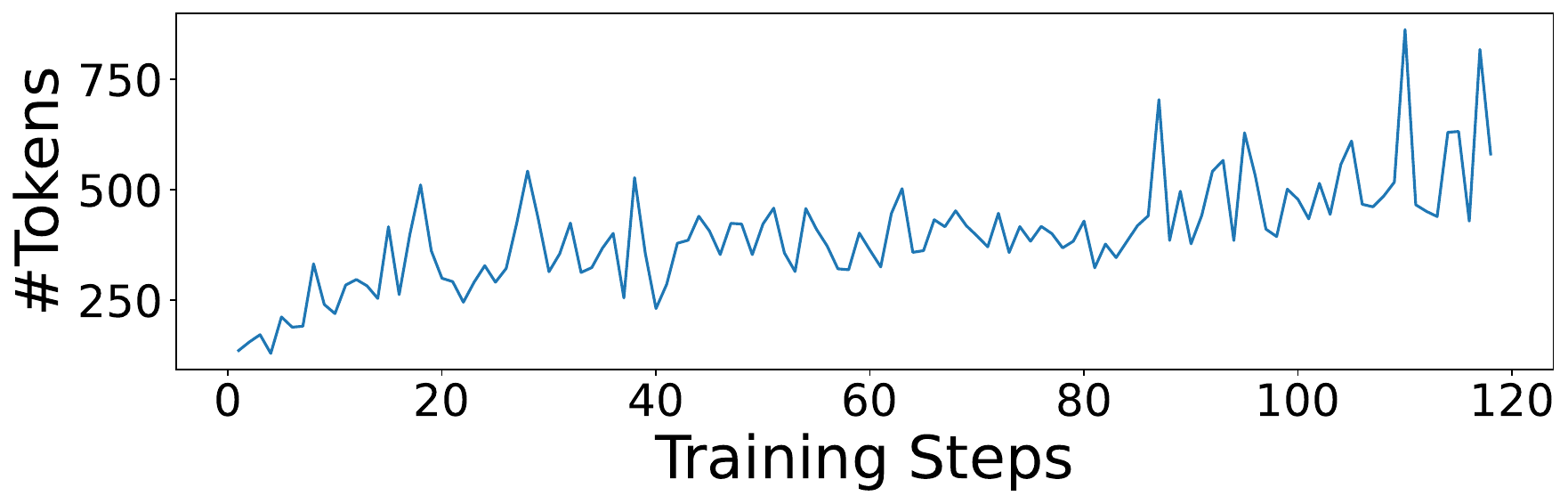}
        \caption{Responses}
    \end{subfigure}
    \caption{Llama-3.1-8B-RLMT-Zero thinks and answers longer as training progresses.}
    \label{fig:thought_and_response_length}
\end{figure}

\paragraph{Increased CoT length.} We also find that as training progresses, the model \textbf{learns to think longer} and \textbf{generate longer responses} (Figure~\ref{fig:thought_and_response_length}). 

\subsection{More experiments and discussion} 

In Appendices~\ref{app:reward_hacking} and~\ref{sec:more_results}, we provide extensive additional results and comparisons:
\begin{itemize}[noitemsep,leftmargin=10px]
    \item \textbf{Reward hacking} (Appendix~\ref{app:reward_hacking}): We study whether RLMT is more prone to reward hacking than RLHF, using both a deliberately hackable reward and ``organic'' reward-hacking features. We find that thinking and non-thinking models hack different features: the non-thinking models hack surface-level features like bullet lists, while the thinking models engage in more subtle behaviors like engagement bait.
    \item \textbf{Additional benchmarks} (Appendix~\ref{sub:additional_benchmarks}): We evaluate our models on IFEval, WildBench-G (evaluated against Gemini 2.5 Flash), MATH-500, and ZebraLogic. We find that RLMT preserves or improves performance on MATH-500 (unlike prior preference optimization methods), though thinking models do not substantially outperform non-thinking models on this benchmark.
    \item \textbf{Human evaluation} (Appendix~\ref{sub:human_eval}): We back our results with LM judges with human evaluation, showing that Llama-3.1-8B-RLMT outperforms the non-thinking counterpart as well as Llama-3.1-70B-Instruct on WildBench.
    \item \textbf{Combined RL on verifiable and unverifiable domains} (Appendix~\ref{sub:combined_rl}): We run RL on a 1:1 mixture of chat and math prompts. The combination improves both chat and math performance, and the thinking model outperforms the non-thinking model uniformly.
    \item \textbf{Comparison with a large model prompted to think} (Appendix~\ref{sub:prompted_70b}): We show that RLMT outperforms a large instruction-tuned model (Llama-3.1-70B-Instruct) that is prompted to think before responding.
    \item \textbf{Comparison with work using alternative rewards} (Appendix~\ref{app:concurrent_reward}): We compare RLMT with approaches that rely on BLEU-based rewards~\citep{Chang2025BLEUBERIBI}, checklist-style rewards~\citep{viswanathan2025checklists}, and TPO~\citep{wu2025thinking}. RLMT consistently achieves substantially stronger results. For example, RLMT outperforms TPO by 19 points on average in the zero-shot setting.
    \item \textbf{Scaling behavior} (Appendices~\ref{sub:inference_budget_scaling} and~\ref{sub:data_size_scaling}): We show how RLMT scales with higher thinking token budgets and more training examples. Performance consistently improves with longer thinking (up to 512+ tokens) and more training data (2K to 8K examples).
\end{itemize}


%% file: sections/related_work.tex
\section{Related Work}
\vspace{-0.2em}

\paragraph{Training stages of language models.}
Modern LMs are typically trained in three stages. First, they are pre-trained on large corpora to acquire general language abilities~\citep{vaswani2017attention,devlin2019bert,radford2019language}.
Second, they undergo supervised fine-tuning (SFT) on curated prompt–response pairs~\citep{radford2018improving,taori2023stanford,wang2023how,grattafiori2024llama3,qwen2025qwen25}, which encourages traits like instruction following~\citep{ouyang2022training,wang2022self}.
Finally, models are refined via reinforcement learning to enforce desirable behavior, either through preference optimization~\citep{ouyang2022training,rafailov2023direct,ethayarajh2024kto,meng2024simpo,ahmadian-etal-2024-back} or domain-specific objectives, such as math~\citep{chen-etal-2024-step,kazemnejad2025vineppo} and tool use~\citep{luo-etal-2025-self}. In this work, we show that RLMT, when applied directly to base models, can also yield strongly aligned models.

\vspace{-0.5em}
\paragraph{Reinforcement learning from human feedback (RLHF).}
To capture subjective attributes such as human preference, RLHF relies on a learned reward model trained on pairwise human judgments~\citep{christiano2017deep,ziegler2020finetuning,ouyang2022training}.
Building on this foundation, preference optimization methods such as DPO~\citep{rafailov2023direct}, KTO~\citep{ethayarajh2024kto}, and SimPO~\citep{meng2024simpo} directly optimize models to generate preferred responses.
Most approaches directly generate the response.
We instead first generate internal reasoning and only later the final answer. Closely related to our work are the efforts that combine preference optimization with chain-of-thought reasoning (for example,~\citet{pang2025bolt,wu2025thinking}). Unlike these approaches, which elicit long CoT traces by prompting instruct models and typically rely on offline algorithms, we show the promise of directly applying online RL with long CoT to base models.

\vspace{-0.5em}
\paragraph{Reinforcement learning with verifiable rewards (RLVR).}
Particularly for domains with objectively verifiable solutions, more suitable algorithms have been proposed, such as GRPO~\citep{shao2024deepseekmath}.
GRPO and its variants~\citep{zheng2025groupsequencepolicyoptimization,yu2025dapo} compute advantages by mean-centering rewards within a group, eliminating the need for a learned critic. DeepSeek-R1~\citep{deepseekai2025deepseekr1} combined GRPO with long CoT reasoning, where models generate reasoning traces that are stripped before evaluation. This enables effective test-time scaling~\citep{snell2025scaling}, and
has also been applied successfully beyond math to other reasoning domains~\citep{liu2025prorl,cheng2025revisiting,huan2025does,ma2025generalreasoner}.
Nonetheless, RLVR remains largely confined to formal settings and has shown limited generalization to open-ended reasoning, the focus of our work.

\vspace{-0.5em}
\paragraph{RLVR beyond rule-based rewards.}
Recent efforts have extended RLVR beyond strict rule-based verification by designing alternative reward signals.
In verifiable domains, reference-free signals such as entropy~\citep{Agarwal2025TheUE} or model confidence~\citep{zhao2025learning} have been shown to be effective. Some other works also explored using compact models to verify responses against ground-truth answers~\citep{ma2025generalreasoner,liu2025noverincentivetraininglanguage} or designing rewards for a specific domain~\citep{gurung2025learning,Wu2025LongWriterZeroMU,Jia2025WritingZeroBT,Li2025SemanticallyAwareRF}.
More closely related to our work, one line explores rubric-based judges~\citep{viswanathan2025checklists,gunjal2025rubrics,wadhwa2025evalagents} or reference-based scores (like BLEU)~\citep{Chang2025BLEUBERIBI} for general chat. These approaches typically do not integrate long CoT reasoning, whereas our work shows that thinking training, when paired with a strong reward model, leads to clear benefits in general chat.

%% file: sections/conclusions.tex
\section{Conclusion}
We have introduced RLMT, which integrates two core components of RLVR - long chain-of-thought and online RL learning, with reward models used in RLHF.
For Llama-3.1-8B and Qwen-2.5-7B (base and instruct), across DPO, PPO, and GRPO, RLMT outperformed standard RLHF by 1.5--3 points on average on a range of benchmarks spanning creative writing and general knowledge.
The models improved by up to 13 points on chat benchmarks, which rivals models orders of magnitude larger, and trained on millions of prompts.
RLMT also showed promise when skipping the SFT warm-start entirely.
Finally, we analyzed the resulting models and found that (1) the RL prompt mixture plays a pivotal role in achieving good performance, and (2) desirable reasoning strategies for open-ended tasks emerge naturally from the training process.
We hope our results inspire future work to enable models to reason for even more general domains.


\section*{Statement on LLM use}
Large language models were used to assist in writing code, and in critiquing and rephrasing the content of the paper. All other original content of this submission originates from humans.

\section*{Acknowledgments}
We thank the anonymous reviewers for their helpful feedback.
We are also grateful to everyone in the Princeton NLP group for support and discussion.
This research is funded by the National Science Foundation (IIS-2211779)
and a Sloan Research Fellowship.

\newpage

%% file: appendices/hyperparameters.tex
\section{Limitations and Future Work}
While our work finds the effectiveness of training LMs with thinking, it is unclear how much of the improvement is due to amplification of traits already present in the model, versus the learning of new traits during the SFT warm-start or RL training~\citep{Yue2025DoesRL,zhu2025surprising}.

The study of this question is important for the design of better training pipelines.
It is also possible that a set of benchmarks larger than the seven considered here would lead to takeaways that were missed here---we choose a reasonable and representative set for the purpose of this paper.
Since our aim was to explore how a simple method can aid model performance, we did not extensively optimize the format used for the internal CoT, the hyperparameters, or the construction of the prompt mixtures.
It is possible that doing so can push our results further, and we invite future work to do so.

\section{Hyperparameters}
\label{app:hyperparameters}

In our warm-start experiments, we used the hyperparameters shown in Table~\ref{tab:sft-hparams} for SFT. 
We provide the corresponding hyperparameters for DPO and PPO/GRPO in Table~\ref{tab:ppo-grpo-hparams}.

We train in bf16 and enable the use of Liger-Kernel~\citep{hsu2025liger} for efficient training.We use the \texttt{trl} library for SFT and DPO~\citep{vonwerra2022trl}, and \texttt{verl} (\url{https://github.com/volcengine/verl}) for PPO and GRPO.
All sampling from the actor for DPO/PPO/GRPO is done at temperature 0.7.

\begin{table}[H]
\centering
\small
\begin{tabular}{lc}
\hline
Hyperparameter & Value \\
\hline
Number of datapoints & 6003 \\
Batch size & 16 \\
Num. epochs & 2 \\
Learning rate & 4e-6 \\
Warmup ratio & 0.1 \\
LR scheduler & cosine \\
Weight decay & 1e-4 \\
Adam betas & (0.9, 0.95) \\
\hline
\end{tabular}
\caption{Supervised fine-tuning (SFT) hyperparameters used in our experiments.}
\label{tab:sft-hparams}
\end{table}

\begin{table}[H]
\centering
\small
\begin{minipage}[t]{0.44\textwidth}
\centering
\begin{tabular}{lc}
\hline
Hyperparameter & Value \\
\hline
\# Prompts & 7560 \\
\# Responses & 8\footnote{Sampled from the initial model before DPO.} \\
Reward model & {\tiny Skywork-Reward-Llama-3.1-8B-v0.2} \\
Batch size & 128 \\
Num. epochs & 2 \\
Learning rate & 3e-7 \\
Warmup ratio & 0.05 \\
LR scheduler & cosine \\
Weight decay & 1e-4 \\
Adam betas & (0.9, 0.95) \\
DPO $\beta$ & 0.1 \\
\hline
\end{tabular}
\caption{Direct Preference Optimization (DPO).}
\label{tab:dpo-hparams}
\end{minipage}\hfill
\begin{minipage}[t]{0.44\textwidth}
\centering
\begin{tabular}{lc}
\hline
Hyperparameter & Value \\
\hline
\# Prompts & 7560 \\
Batch size & 64 \\
Samples per prompt & 8 \\
Group size & 8$^\text{g}$ \\
Max prompt length & 1024 \\
Max response length & 4096 \\
Num. steps & 120 (1 epoch) \\
Actor learning rate & 1e-6\footnote{We used 3e-7 for the warm-started instruct models.} \\
Critic learning rate & 1e-5$^\text{p}$ \\
Weight decay & 0.01 \\
Scheduler & constant \\
Warmup ratio & 0 \\
Advantage estimator & GRPO/GAE \\
KL coefficient & 0.001 \\
\hline
\end{tabular}
\caption{PPO / GRPO. Entries marked $^\text{p}$ or $^\text{g}$ are only used for PPO or GRPO, respectively.}
\label{tab:ppo-grpo-hparams}
\end{minipage}
\end{table}

%% file: appendices/reward_hacking.tex
\section{Is RLMT more prone to reward hacking than RLHF?}
\label{app:reward_hacking}

In Section~\ref{sec:training}, we found that RLMT is sensitive to the choice of reward model. In this section, we check whether this means that RLMT is more prone to reward hacking than RLHF. In other words, does RL on a thinking model reward hack more than RL on a non-thinking model? To this end, we perform two experiments.

\paragraph{RL with a fake reward.} We repeat the RL training of the Llama-3.1-8B SFT checkpoints from Section~\ref{sec:main_results}, but add half-standard-deviation (=6.81) of reward to the Skywork-RM whenever the response contains the word ``absolutely'' (case-insensitive). Then, we measure the percentage of rollouts containing the word ``absolutely'' over the course of 100 gradient steps. We report this percentage over the course of 100 training steps in Table~\ref{tab:fake-reward-hacking}.

\begin{wraptable}{r}{0.34\textwidth}
    \centering
    \small
    \renewcommand{\tabcolsep}{1.5mm}
    \caption{Percentage of rollouts containing ``absolutely'' during RL with a fake reward.}
    \label{tab:fake-reward-hacking}
    \begingroup
    \begin{tabular}{rrr}
        \toprule
        \textbf{\emph{Step}} & \textbf{\emph{Nothink (\%)}} & \textbf{\emph{Think (\%)}} \\
        \midrule
        0   & \textbf{3}  & 1  \\
        40  & 3  & \textbf{15} \\
        70  & 19 & \textbf{96} \\
        100 & 95 & \textbf{96} \\
        \bottomrule
    \end{tabular}
    \endgroup
\end{wraptable}

Based on this result (Table~\ref{tab:fake-reward-hacking}), it does appear that thinking models exploit a hackable reward sooner - though both types of models eventually converge to near-maximum hacking.

\paragraph{Organic signs of reward hacking.} We also check the two GRPO checkpoints atop Llama-3.1-8B (non-thinking and thinking, respectively) for ``organic'' signs of reward hacking that have been widely discussed in popular media. We measure four such features across model responses to the prompts in AlpacaEval2 and WildBench: (1) length (\#words), (2) bullet lists (avg. \#bullets per response), (3) engagement-bait closer (0/1 score - for example, ``Would you like hear about some ways you can do X?''), and (4) Self-praise (e.g., ``Here is a comprehensive response'', ``I have compiled a thorough report'') as a 0/1 score. We measure each of the four attributes as the difference between the SFT and RL checkpoints, i.e., RL - SFT, averaged over all the prompts. We report these results in Table~\ref{tab:organic-reward-hacking}.

\begin{table}[h]
    \centering
    \small
    \renewcommand{\tabcolsep}{1.5mm}
    \caption{Changes in organic reward-hacking features from SFT to RL. Length is measured in words, bullets as counts, and engagement-bait and self-praise as rates. Reported numbers are averaged over all prompts in AlpacaEval2 and WildBench.}
    \label{tab:organic-reward-hacking}
    \begingroup
    \begin{tabular}{lrrrl}
        \toprule
        \textbf{\emph{Feature (unit)}} & \textbf{\emph{Nothink}} & \textbf{\emph{Think}} & \textbf{\emph{Think $-$ Nothink}} & \textbf{\emph{More}} \\
        \midrule
        Length (words) & \textbf{+166} & \textbf{+170} & +3.6 & tie \\
        Bullets (count) & \textbf{+6.83} & +3.10 & -3.73 & nothink \\
        Engagement-bait closer (rate) & +0.176 & \textbf{+0.232} & +0.056 & think \\
        Self-praise (rate) & +0.114 & \textbf{+0.217} & +0.103 & think \\
        \bottomrule
    \end{tabular}
    \endgroup
\end{table}

Our results (Table~\ref{tab:organic-reward-hacking}) suggest that thinking models do not necessarily exploit length more, and use fewer bullet lists than non-thinking models. On the other hand, more subtle kinds of reward hacking like engagement bait (that have been observed with recent GPT and Claude models), show an increase. 
In effect, reward hacking remains an important problem for both thinking and non-thinking models.

%% file: appendices/more_results.tex
\section{Additional Results}
\label{sec:more_results}

We present in this appendix results on benchmarks beyond those used in the main text.

\subsection{Benchmark descriptions and additional evaluation}
\label{sub:additional_benchmarks}

We evaluate the models trained in this paper on the following benchmarks:
\begin{enumerate}
    \item \textbf{[WB] WildBench}~\citep{lin2025wildbench} evaluates models on their ability to converse with users. It has 1,024 user prompts, some of which involve multiple turns. Unlike AlpacaEval2, WildBench is evaluated using an instance-wise manually checked rubric which is less susceptible to reward hacking. Each response is compared against a reference response from GPT-4, and scored one of -100 (much worse), -50 (worse), 0 (similar), 50 (better), or 100 (much better). The final score is the mean of the instance-wise scores.
    \item \textbf{[AE2] AlpacaEval2}~\citep{dubois2024lengthcontrolled} has 805 user prompts paired with reference responses from GPT-4-1106-preview. It outputs a head-to-head win rate between 0--100\% as generated by a generative judge. We use the length-controlled win-rate as recommended by~\citet{dubois2024lengthcontrolled}, but replace the default GPT-4 judge with GPT-4o.
    \item \textbf{[AH2] ArenaHardV2}~\citep{li2024from,li2024fromb} has 500 challenging real-world user queries. We use the setting with a GPT-4.1 judge and style control to mitigate potential bias.    
    \item \textbf{[CW3] Creative Writing V3}~\citep{samuel2025creativewritingv3} evaluates models on their ability to write 96 story chapters under various constraints. We generate an absolute score between 0--100 using GPT-4.1 as the judge.
    \item \textbf{PopQA}~\citep{mallen2022when} consists of around 14k factual questions about popular and less known entities. The result is a percentage score between 0--100\%.
    \item \textbf{[IF\textsubscript{Ben}] IFBench}~\citep{pyatkin2025generalizing} provides models with 294 prompts from varied domains that each has multiple constraints such as ``include three numbers in your 22-nd sentence.'' We average the compliance rate across all examples to produce a score between 0--100.
    \item \textbf{[MMLU\textsubscript{R}] MMLU-Redux}~\citep{gema2025are} is a manually cleaned version of the MMLU~\citep{hendrycks2021measuring} benchmark, consisting of 5,700 questions that test the model's ability to answer general knowledge questions across 57 subjects.
    \item \textbf{IFEval, \textit{only in Appendix~\ref{sec:more_results}}},~\citep{zhou2023instruction} provides models with 541 simple questions under a set of constraints such as ``do not use commas,'' and generates a score between 0--100 that signifies how well the model follows instructions. This is an alternate evaluation of instruction-following capabilities.
    \item \textbf{WildBench v.s. Gemini 2.5 Flash Preview 0520 (WildBench-G, \textit{only in Appendix~\ref{sec:more_results}}}),~\citep{lin2025wildbench} is WildBench, but evaluated against reference responses from Gemini 2.5 Flash Preview 0520. Our motivation for including this is that beyond a certain number, being X\% better than GPT-4-Turbo's responses (the default reference) may be less correlated with actual improvements.
    \item \textbf{MATH-500, \textit{only in Appendix~\ref{sec:more_results}}},~\citep{lightman2023lets} consists of 500 hard math problems filtered from the MATH dataset~\citep{hendrycks2021measuring} by OpenAI. We report the exact match accuracy from 0--100\%.
    \item \textbf{ZebraLogic, \textit{only in Appendix~\ref{sec:more_results}}},~\citep{lin2025zebralogic} tests language models on 1,000 logical grid puzzles. We report the exact match accuracy from 0--100\%.
\end{enumerate}

We list the corresponding results in Table~\ref{tab:more_results}. We observe that
\begin{enumerate}
    \item Neither the thinking nor non-thinking models perform well on IFEval. Therefore we attribute blame here to the reward model: a preference model finds it difficult to determine if specific instructions like ``do not use commas'' are followed. Lacking a reliable reward model, GRPO is (and other objectives are) not able to optimize for the desired behavior.
    \item We see large gaps on WildBench-G, consistent with our findings in the main text. In fact, the gaps here are larger by 0.5--1 points; the difference in quality is more important when the reference is better.
    \item While prior work optimizing user preferences~\citep{rafailov2023direct,meng2024simpo} found that it tanked scores on MATH-500, we find that our post-training actually improves mathematical abilities compared to the SFT models. On the other hand, the thinking and non-thinking models perform similarly (and the latter sometimes even outperforms the former)---therefore, building a thinking post-training pipeline for all domains constitutes a useful direction for future work.
    \item While the initial models perform poorly on ZebraLogic, our thinking post-training shaves a further 1--3 points off the scores.
\end{enumerate}

\begin{table}[h]
\centering
\small
\renewcommand{\tabcolsep}{1.2mm}
\caption{Additional benchmark results for all model variants. $\faLightbulbO$ indicates long chain-of-thought. Best numbers are \textbf{bold} and second best are \underline{underlined} within each model group.}
    \label{tab:more_results}
    \begingroup
\begin{tabular}{lllrrrrr}
\toprule
\textbf{\emph{Backbone}} & \textbf{\emph{Objective}} & \textbf{\emph{\faLightbulbO }} & \textbf{\emph{IFEval}} & \textbf{\emph{WB-G}} & \textbf{\emph{MATH-500}} & \textbf{\emph{Zebra}} \\
\cmidrule{1-7}
Llama-3.1-8B & + SFT & $\times$ & 50.5 & -63.3 & 13.0 & \textbf{8.4} \\
\rowcolor{gray!14}
 &  & $\checkmark$ & \textbf{56.0} & -61.3 & 18.8 & 6.2 \\
 & \quad     + DPO & $\times$ & \underline{51.4} & -51.8 & 15.6 & 7.6 \\
\rowcolor{gray!14}
 &  & $\checkmark$ & 45.1 & -50.5 & \textbf{21.2} & \underline{7.7} \\
 & \quad     + PPO   & $\times$ & 42.0 & -51.8 & 17.6 & 7.2 \\
\rowcolor{gray!14}
 &  & $\checkmark$ & 43.4 & -51.4 & \underline{20.4} & 7.3 \\
 & \quad     + GRPO & $\times$ & 36.8 & \underline{-34.4} & 18.8 & 5.9 \\
\rowcolor{gray!14}
 &  & $\checkmark$ & 36.4 & \textbf{-27.1} & 19.4 & 7.0 \\
\cmidrule{1-7}
Qwen2.5-7B & + SFT & $\times$ & 46.6 & -58.8 & 60.8 & \underline{8.7} \\
\rowcolor{gray!14}
 &  & $\checkmark$ & \underline{56.4} & -55.0 & 63.0 & 7.8 \\
 & \quad     + DPO & $\times$ & 44.7 & -56.2 & 63.2 & \textbf{9.0} \\
\rowcolor{gray!14}
 &  & $\checkmark$ & \textbf{57.1} & -47.5 & 64.8 & 7.7 \\
 & \quad     + PPO   & $\times$ & 46.0 & \underline{-45.4} & 61.6 & 8.2 \\
\rowcolor{gray!14}
 &  & $\checkmark$ & 50.3 & -45.5 & \textbf{66.0} & 7.9 \\
 & \quad     + GRPO & $\times$ & 38.4 & -46.4 & \underline{65.0} & 7.3 \\
\rowcolor{gray!14}
 &  & $\checkmark$ & 46.8 & \textbf{-33.3} & 64.4 & 6.7 \\
\cmidrule{1-7}
Llama-3.1-8B-Instruct &  & $\times$ & \textbf{75.6} & -65.8 & 45.2 & \textbf{13.3} \\
 & + SFT & $\times$ & 70.8 & -53.3 & 28.8 & 10.3 \\
\rowcolor{gray!14}
 &  & $\checkmark$ & \underline{72.3} & -51.4 & 40.6 & 11.0 \\
 & \quad     + DPO & $\times$ & 71.2 & -44.2 & 36.2 & 11.0 \\
\rowcolor{gray!14}
 &  & $\checkmark$ & 69.3 & -42.3 & 43.0 & \underline{11.6} \\
 & \quad     + PPO   & $\times$ & 62.5 & -27.1 & 40.6 & 10.8 \\
\rowcolor{gray!14}
 &  & $\checkmark$ & 54.9 & \underline{-18.6} & 41.2 & 10.9 \\
 & \quad     + GRPO & $\times$ & 61.2 & -27.5 & \textbf{50.6} & 8.8 \\
\rowcolor{gray!14}
 &  & $\checkmark$ & 57.1 & \textbf{-15.8} & \underline{45.2} & 10.8 \\
\cmidrule{1-7}
Qwen2.5-7B-Instruct &  & $\times$ & \textbf{71.5} & -59.2 & \textbf{74.6} & \underline{10.3} \\
 & + SFT & $\times$ & 61.6 & -56.2 & 54.0 & 7.5 \\
\rowcolor{gray!14}
 &  & $\checkmark$ & \underline{66.4} & -52.4 & 62.6 & 8.9 \\
 & \quad     + DPO & $\times$ & 62.1 & -53.3 & 64.4 & 8.8 \\
\rowcolor{gray!14}
 &  & $\checkmark$ & 65.1 & -46.8 & 63.0 & \underline{9.3} \\
 & \quad     + PPO   & $\times$ & 63.8 & -42.7 & 64.6 & 9.2 \\
\rowcolor{gray!14}
 &  & $\checkmark$ & 65.4 & \underline{-35.6} & 63.4 & 8.7 \\
 & \quad     + GRPO & $\times$ & 63.0 & -38.9 & \underline{65.8} & 8.5 \\
\rowcolor{gray!14}
 &  & $\checkmark$ & 60.1 & \textbf{-31.6} & 65.0 & 7.4 \\
\bottomrule
\end{tabular}
\endgroup
\end{table}

\subsection{Human evaluation}
\label{sub:human_eval}

In this section, we account for possible judge bias in the automated evaluation metrics by performing human evaluation.

We selected 50 random prompts from WildBench and then selected one response each from the three models - (1) Llama-3.1-8B-SFT + GRPO (thinking, RLMT), (2) Llama-3.1-8B-SFT + GRPO (no thinking), and (3) Llama-3.1-70B-Instruct. We abbreviate these models to ``8B-NT'', ``8B-T'', and ``70B-I'', respectively. We then ranked each set of three responses in a blind human evaluation. Table~\ref{tab:human-evaluation} reports the \textbf{Mean Reciprocal Rank (MRR, higher is better)} and \textbf{pairwise win-rates (\%)} of the three models.

\begin{wraptable}{r}{0.48\textwidth}
    \centering
    \renewcommand{\tabcolsep}{1.0mm}
    \caption{Blind human evaluation over 50 WildBench prompts. We report mean reciprocal rank (MRR) and pairwise win rates (WR).}
    \label{tab:human-evaluation}
    \begingroup
    \begin{tabular}{lrrrr}
        \toprule
        \textbf{\emph{Model}} & \textbf{\emph{MRR}} & \multicolumn{3}{c}{\textbf{\emph{WR vs. (\%)}}} \\
        \cmidrule(lr){3-5}
          & & \textbf{\emph{8B-T}} & \textbf{\emph{8B-NT}} & \textbf{\emph{70B-I}} \\
        \midrule
        8B-T  & \textbf{0.72} & --   & \textbf{62.0} & \textbf{60.0} \\
        8B-NT & 0.55 & 38.0 & --   & 40.0 \\
        70B-I & 0.57 & \textbf{40.0} & 60.0 & --   \\
        \bottomrule
    \end{tabular}
    \endgroup
\end{wraptable}

We note that 8B-T obtains a distinctly higher MRR than both 8B-NT and 70B-I, and has a win rate of 60.0\% or above against both, individually (Table~\ref{tab:human-evaluation}).

Qualitatively, we noticed that 70B-I produced very short responses - e.g., listing a plain bullet list with 6-8 words of description when asked for places to visit (8B-NT and 8B-T both provided a lot of detail, in contrast). This was in fact preferred in some questions regarding, e.g., math problems; but overall, it hurt 70B-I's performance. We did not observe a significant difference between the output lengths of 8B-T and 8B-NT.

\clearpage
\subsection{Combined RL on unverifiable and verifiable domains}
\label{sub:combined_rl}

In this section, we test whether the combination of RL on verifiable domains (RLVR) and unverifiable domains (RLMT) can improve performance on both domains.

\begin{wraptable}[16]{r}{0.5\textwidth}
    \vspace{-1em}
    \centering
    \scriptsize
    \renewcommand{\tabcolsep}{1.2mm}
    \caption{Results from RL on chat prompts, math prompts, or a 1:1 mixture of both domains.}
    \label{tab:combined-rlmt-rlvr}
    \begingroup
    \begin{tabular}{lrrr}
        \toprule
        \textbf{\emph{Model}} & \textbf{\emph{MATH-500 (\%)}} & \textbf{\emph{AE2 (\%)}} & \textbf{\emph{WB (score)}} \\
        \midrule
        SFT & & & \\
        \quad nothink & 13.0 & 26.8 & -10.6 \\
        \quad think   & 18.8 & 29.7 & -1.6 \\
        \midrule
        RL, chat only & & & \\
        \quad nothink & 18.8 & 46.7 & 33.2 \\
        \quad think   & 19.4 & \textbf{52.3} & \textbf{38.1} \\
        \midrule
        RL, chat:math = 1:1 & & & \\
        \quad nothink & 21.4 & 36.6 & 28.8 \\
        \quad think   & 24.0 & 39.0 & 33.2 \\
        \midrule
        RL, math only & & & \\
        \quad nothink & 21.5 & 25.3 & -7.4 \\
        \quad think   & \textbf{24.7} & 28.5 & -1.8 \\
        \bottomrule
    \end{tabular}
    \endgroup
\end{wraptable}

To test the combination, we performed the following experiment on top of the two SFT checkpoints from Llama-3.1-8B (thinking and non-thinking, see top row of Table~\ref{tab:grpo-benchmarks}). We filtered those prompts from the DeepScaleR-preview dataset\footnote{\url{https://huggingface.co/datasets/agentica-org/DeepScaleR-Preview-Dataset}} where the pass rate of either model was between 20\% and 80\% over 16 responses. This leads to 7144 chat examples (from our original setting), and 8172 math examples to perform RL on. We check three settings: (1) chat-only RL (RLMT) (2) chat-math RL where we select each only one domain per-batch (i.e., alternate between the two) (3) math-only RL (RLVR).

We perform all three experiments with the same hyperparameters as the submission, and for 120 gradient steps.

We observe from Table~\ref{tab:combined-rlmt-rlvr}:
\begin{enumerate}
    \item \textit{RLVR+RLMT can improve both chat and math performance.} When combining chat and math prompts, we achieve performance on MATH-500 close to that of RLVR on math only. Conversely, we also achieve a strong performance on the two chat benchmarks (+22 points on average), though this performance is a bit worse than that of pure RLMT.
    \item \textit{The thinking model outperforms the non-thinking model on all three benchmarks.} In the combined RL setting, we see a gap of 2.5-4 points across all three benchmarks.
\end{enumerate}

\subsection{Comparison with large model prompted to think}
\label{sub:prompted_70b}

A possible explanation of our results can be that they are a result of the thinking process and not the training of RLMT itself, which can be performed by an instruction-tuned model. Below, we copy the results of the two Llama-3.1-8B (Base) GRPO checkpoints (think and no-think), and the original Llama-3.1-70B-Instruct results. We also include a new row, where we prompt Llama-3.1-70B-Instruct to think before responding, with the following system prompt:

\begin{quote}
\texttt{You are a helpful assistant. When prompted, first plan your response using <think>...</think> tags. Then, respond to the user's question. That is, use the following format when responding:}

\texttt{<think> [thought process and planning] </think>}

\texttt{[response]}
\end{quote}

We show the results in Table~\ref{tab:prompted-70b}.

\begin{table}[h]
    \centering
    \small
    \renewcommand{\tabcolsep}{1.5mm}
    \caption{Comparison of Llama-3.1-8B GRPO checkpoints against Llama-3.1-70B-Instruct with and without a thinking prompt.}
    \label{tab:prompted-70b}
    \begingroup
    \begin{tabular}{lrrrr}
        \toprule
        \textbf{\emph{Model}} & \textbf{\emph{AE2 (\%)}} & \textbf{\emph{WB (score)}} & \textbf{\emph{CW3 (score)}} & \textbf{\emph{Avg. (score)}} \\
        \midrule
        Llama-3.1-8B + SFT + GRPO (nothink) & 46.7 & 33.2 & \textbf{84.2} & 54.7 \\
        Llama-3.1-8B + SFT + GRPO (think) & \textbf{52.3} & \textbf{38.1} & 80.9 & \textbf{57.1} \\
        Llama-3.1-70B-Instruct & & & & \\
        \quad nothink-prompt & 42.0 & 16.3 & 59.4 & 39.2 \\
        \quad think-prompt & 21.8 & -14.5 & 54.8 & 20.7 \\
        \bottomrule
    \end{tabular}
    \endgroup
\end{table}

We find that the 70B instruct model performs \textit{worse} when prompted to think first (Table~\ref{tab:prompted-70b}). The takeaway remains that the 8B thinking model outperforms the other models handily; and thus that the GRPO training is an integral part of the RLMT method.

\begin{table}
    \small
    \centering
    \caption{Comparison of RLMT models with concurrent work using different reward signals for RLVR }
    \label{tab:otherreward-vs-ours}
    \begingroup
    \begin{tabular}{llrrrrr}
        \toprule
        \textbf{\emph{Model}} & \textbf{\emph{Backbone}} & \textbf{\emph{Avg.}} & \textbf{\emph{WB}} & \textbf{\emph{AE2}} & \textbf{\emph{AH2}} & \textbf{\emph{CWv3}} \\
        \midrule
        \multicolumn{6}{c}{\textit{(Models sharing similar backbones)}} \\
        \multicolumn{5}{l}{\textit{\textbf{Our models}}} \\
        Llama-3.1-8B-RLMT-Zero & Llama-3.1-8B (Base)& 24.0 & 7.2 & 34.0 & 5.6 & 49.0\\
        Qwen2.5-7B-RLMT-Zero &  Qwen-2.5-7B (Base) & 30.2 & 22.2 & 54.0 & 10.8 & 54.0\\
        \rowcolor{green!14} Llama-3.1-8B-Inst-RLMT & Llama-3.1-8B-Instruct&  \textbf{54.1} & \textbf{50.4} & \textbf{58.7} & 22.9 & \textbf{84.3}\\
        Qwen2.5-7B-Inst-RLMT &  Qwen-2.5-7B-Instruct &48.3 & 46.3 & 50.5 & 20.8 & 75.6\\
        \multicolumn{5}{l}{\textit{\textbf{Models trained using BLEU-based reward or checklist-based reward}}} \\
        Llama3.1-8B-BLEUBERI  & Llama-3.1-8B (Base)& 11.6 & -18.5 & 17.4 & 2.4& 45.3 \\
        Qwen2.5-7B-BLEUBERI &  Qwen-2.5-7B (Base) & 19.7 & -6.5 & 29.5 & 12.8 &  42.8\\
        Qwen2.5-7B-RLCF &   Qwen-2.5-7B-Instruct & 32.1 & 23.6 & 40.2 & 10.4 & 54.3\\
        \bottomrule
    \end{tabular}
    \endgroup
\end{table}

\subsection{Comparison with concurrent RLVR work using alternative rewards}
\label{app:concurrent_reward}
Our work extends RLVR to general domains by leveraging a reward model. Concurrent efforts have also sought to move beyond verifiable domains through alternative reward designs, such as BLEU-based reward to reference responses~\citep{Chang2025BLEUBERIBI} and checklist-style rewards where an LM scores outputs against LM-generated\\ rubrics~\citep{viswanathan2025checklists}. We compare RLMT against these open-source models of these concurrent work on open-ended benchmarks (chat and creative writing). Specifically, we include: (1) Llama-3.1-8B-BLEUBERI, (2) Qwen-2.5-7B-BLEUBERI, and (3) Qwen-2.5-7B-RLCF. Table~\ref{tab:otherreward-vs-ours} summarizes these models along with their source backbones.

Across the same backbones, RLMT consistently achieves substantially stronger results. For example, Llama-3.1-8B-RLMT-Zero outperforms Llama-3.1-8B-BLEUBERI by roughly 13 points on average across open-ended benchmarks, despite the latter leveraging reference responses. These findings suggest that a strong reward model provides robust and effective signal for online RL training in general-purpose domains, surpassing these alternative reward designs.

\begin{wraptable}{r}{0.5\textwidth}
    \small
    \centering
    \caption{Comparison of RLMT with TPO on chat benchmarks, on Llama-3.1-8B.}
    \label{tab:tpo-vs-rlmt}
    \begingroup
    \begin{tabular}{lrrrr}
        \toprule
        \textbf{\emph{Setting}} & \textbf{\emph{WB}} & \textbf{\emph{AE2}} & \textbf{\emph{AH2}} & \textbf{\emph{ChatAvg}} \\
        \midrule
        Warm-start & & & & \\
        \quad +RLMT & \
        \textbf{38.1} & \textbf{52.3} & \textbf{15.9} & \textbf{35.4} \\
        \quad +TPO & 23.7 & 43.5 & 13.9 & 27.0 \\
        \midrule
        Zero & & & & \\
        \quad +RLMT & 7.2 & 34.0 & 5.6 & 15.6 \\
        \quad +TPO & -31.2 & 15.2 & 2.2 & -4.6 \\
        \bottomrule
    \end{tabular}
    \endgroup
\end{wraptable}

In addition, we also compare RLMT with TPO~\citep{wu2025thinking}. Table~\ref{tab:tpo-vs-rlmt} shows the comparison on chat benchmarks.
We see that RLMT substantially outperforms TPO on chat and creative writing---especially in the ``zero'' setting, where the difference is 19 points on average.

\subsection{Scaling behavior of RLMT}

We investigate how RLMT performance changes when scaling inference budgets (thinking token length) and training data size. Due to compute constraints, we cannot scale model sizes (scaling to 12B or larger models would require 16 or more GPUs with 80 GB of memory each).

\subsubsection{Inference budget scaling}
\label{sub:inference_budget_scaling}

We evaluate our RLMT model (Llama-3.1-8B-Warm-start-RLMT) under different thinking-token budgets by comparing three settings: (1) forcing a maximum of 127 thinking tokens (by closing thinking with \texttt{</think>} early and forcing the model to answer), (2) our main setting, which does not constrain thinking length and yields an average of $\sim$260 thinking tokens across benchmarks, and (3) encouraging additional thinking by appending ``Wait, let's re-analyze the plan to make a better response'' whenever the model produces fewer than 513 thinking tokens. As shown in Table~\ref{tab:inference-budget}, restricting the model to fewer thinking tokens causes substantial drops in performance. Interestingly, encouraging the model to think more yields moderate improvements on chat benchmarks.

\begin{table}[h]
    \small
    \centering
    \caption{Performance of RLMT under different thinking-token budgets. Performance consistently improves with more thinking tokens.}
    \label{tab:inference-budget}
    \begingroup
    \renewcommand{\tabcolsep}{1.2mm}
    \begin{tabular}{lrrrrrrrrr}
        \toprule
        \textbf{\emph{Budget}} & \textbf{\emph{WB}} & \textbf{\emph{AE2}} & \textbf{\emph{AH2}} & \textbf{\emph{ChatAvg}} & \textbf{\emph{CWv3}} & \textbf{\emph{PopQA}} & \textbf{\emph{IF\textsubscript{Ben}}} & \textbf{\emph{MMLU\textsubscript{R}}} & \textbf{\emph{OverallAvg}} \\
        \midrule
        $<$ 128 & 23.5 & 44.9 & 13.1 & 27.2 & 73.3 & 30.2 & 17.6 & 56.3 & 37.0 \\
        $\sim$260 (Main) & 38.1 & 52.3 & 15.9 & 35.4 & 80.9 & 30.3 & 15.6 & \textbf{61.7} & 42.1 \\
        $>$ 512 & \textbf{42.7} & \textbf{52.6} & \textbf{17.2} & \textbf{37.5} & \textbf{82.7} & \textbf{30.4} & \textbf{16.3} & 60.5 & \textbf{43.2} \\
        \bottomrule
    \end{tabular}
    \endgroup
\end{table}

\subsubsection{Data size scaling}
\label{sub:data_size_scaling}

We also conduct a scaling study by varying the amount of RL training data (2K, 4K, 8K) using the WildChat-IF dataset (which provides up to 8K examples). As shown in Table~\ref{tab:data-scaling}, using more training data consistently improves performance, especially on chat benchmarks.

\begin{table}[h]
    \small
    \centering
    \caption{Performance of RLMT with varying amounts of training data. Performance consistently improves with more training data.}
    \label{tab:data-scaling}
    \begingroup
    \renewcommand{\tabcolsep}{1.2mm}
    \begin{tabular}{lrrrrrrrrr}
        \toprule
        \textbf{\emph{Setting}} & \textbf{\emph{WB}} & \textbf{\emph{AE2}} & \textbf{\emph{AH2}} & \textbf{\emph{ChatAvg}} & \textbf{\emph{CWv3}} & \textbf{\emph{PopQA}} & \textbf{\emph{IF\textsubscript{Ben}}} & \textbf{\emph{MMLU\textsubscript{R}}} & \textbf{\emph{OverallAvg}} \\
        \midrule
        WildChat 2K & 37.7 & 44.1 & 14.4 & 32.1 & 80.9 & 30.2 & \textbf{16.7} & 57.7 & 40.2 \\
        WildChat 4K & 37.4 & 48.5 & 15.5 & 33.8 & \textbf{81.2} & 29.8 & 16.3 & 61.5 & 41.5 \\
        WildChat 8K (Main) & \textbf{38.1} & \textbf{52.3} & \textbf{15.9} & \textbf{35.4} & 80.9 & \textbf{30.3} & 15.6 & \textbf{61.7} & \textbf{42.1} \\
        \bottomrule
    \end{tabular}
    \endgroup
\end{table}

\subsubsection{GRPO achieves the best performance, but RLMT remains effective with DPO and PPO.}
\label{sub:ppo}

\begin{table}[t]
    \centering
    \small
    \renewcommand{\tabcolsep}{1.0mm}
    \fontsize{7.5}{7.5}\selectfont
    \vspace{-1em}
    \caption{DPO/PPO results for warm-start and zero training. \faLightbulbO\ shows whether thinking is enabled, with $\checkmark$ denoting RLMT models and $\times$ denoting RLHF models. The best numbers are \textbf{bolded} and the second are \underline{underlined}. Warm-start + RLMT remains effective with DPO/PPO, but they lag GRPO. The two algorithms are ineffective compared to GRPO for zero training.
    }
        \vspace{-0.75em}
        \label{tab:dpo-ppo-benchmarks}
        \begingroup
    \begin{tabular}{lllrrrrrrrrrrr}
    \toprule
    \textbf{\emph{Backbone}} & \textbf{\emph{Training}} & \textbf{\emph{\faLightbulbO }} & \textbf{\emph{WB}} & \textbf{\emph{AE2}} & \textbf{\emph{AH2}} & \textbf{\emph{Avg\textsubscript{Chat}}} & \textbf{\emph{}} & \textbf{\emph{CWv3}} & \textbf{\emph{PopQA}} & \textbf{\emph{IF\textsubscript{Ben}}} & \textbf{\emph{MMLU\textsubscript{R}}} & \textbf{\emph{}} & \textbf{\emph{Avg}} \\
    \cmidrule{1-7} \cmidrule{9-12} \cmidrule{14-14}
    \multicolumn{14}{l}{\underline{\textit{SFT Warm-Started Models}}} \\
    Llama-3.1-8B & + SFT & $\times$ & \cellcolor{blue!14}-10.6 & \cellcolor{blue!14}26.8 & \cellcolor{blue!14}5.1 & \cellcolor{blue!14}7.1 &  & 75.2 & 26.4 & 15.6 & 59.7 &  & 28.3 \\
    \rowcolor{gray!14}
      &  & $\checkmark$ & \cellcolor{blue!14}-1.6 & \cellcolor{blue!14}29.7 & \cellcolor{blue!14}6.5 & \cellcolor{blue!14}11.5 &  & 75.1 & 30.5 & 17.0 & 61.1 &  & 31.2 \\
      &   + DPO & $\times$ & \cellcolor{blue!14}14.4 & \cellcolor{blue!14}34.6 & \cellcolor{blue!14}9.0 & \cellcolor{blue!14}19.3 &  & \underline{78.3} & 28.2 & 17.0 & 60.3 &  & 34.5 \\
    \rowcolor{gray!14}
      &  & $\checkmark$ & \cellcolor{blue!14}17.3 & \cellcolor{blue!14}36.8 & \cellcolor{blue!14}\underline{10.9} & \cellcolor{blue!14}\underline{21.7} &  & 76.6 & \underline{32.4} & \underline{17.3} & \underline{62.5} &  & 36.3 \\
      &   + PPO   & $\times$ & \cellcolor{blue!14}\textbf{23.4} & \cellcolor{blue!14}\underline{40.2} & \cellcolor{blue!14}\textbf{11.9} & \cellcolor{blue!14}\textbf{25.2} &  & 71.3 & 29.6 & 16.7 & 62.1 &  & \underline{36.5} \\
    \rowcolor{gray!14}
      &  & $\checkmark$ & \cellcolor{blue!14}\underline{21.7} & \cellcolor{blue!14}\textbf{43.3} & \cellcolor{blue!14}10.7 & \cellcolor{blue!14}\textbf{25.2} &  & \textbf{81.9} & \textbf{33.0} & \textbf{18.0} & \textbf{64.3} &  & \textbf{39.0} \\
    \cmidrule{1-7} \cmidrule{9-12} \cmidrule{14-14}
    Qwen2.5-7B & + SFT & $\times$ & \cellcolor{blue!14}-0.9 & \cellcolor{blue!14}28.8 & \cellcolor{blue!14}8.0 & \cellcolor{blue!14}12.0 &  & 61.9 & 21.1 & \underline{19.0} & 65.2 &  & 29.0 \\
    \rowcolor{gray!14}
      &  & $\checkmark$ & \cellcolor{blue!14}12.0 & \cellcolor{blue!14}33.9 & \cellcolor{blue!14}10.6 & \cellcolor{blue!14}18.8 &  & 69.5 & 22.0 & \textbf{21.1} & 62.1 &  & 33.0 \\
      &   + DPO & $\times$ & \cellcolor{blue!14}12.8 & \cellcolor{blue!14}32.5 & \cellcolor{blue!14}11.6 & \cellcolor{blue!14}19.0 &  & 70.5 & 21.4 & 17.0 & \textbf{70.1} &  & 33.7 \\
    \rowcolor{gray!14}
      &  & $\checkmark$ & \cellcolor{blue!14}24.9 & \cellcolor{blue!14}38.1 & \cellcolor{blue!14}\underline{14.1} & \cellcolor{blue!14}\underline{25.7} &  & 74.8 & \textbf{22.6} & \textbf{21.1} & 66.4 &  & \underline{37.4} \\
      &   + PPO   & $\times$ & \cellcolor{blue!14}\underline{29.0} & \cellcolor{blue!14}35.5 & \cellcolor{blue!14}\underline{14.9} & \cellcolor{blue!14}\underline{26.5} &  & 69.7 & 21.9 & 18.0 & \underline{68.6} &  & 36.8 \\
    \rowcolor{gray!14}
      &  & $\checkmark$ & \cellcolor{blue!14}\textbf{30.9} & \cellcolor{blue!14}\textbf{42.2} & \cellcolor{blue!14}\textbf{15.2} & \cellcolor{blue!14}\textbf{29.4} &  & \textbf{77.2} & \underline{22.3} & \underline{20.4} & 67.1 &  & \textbf{39.3} \\
    \cmidrule{1-7} \cmidrule{9-12} \cmidrule{14-14}
    Llama-3.1-8B-Instruct &  & $\times$ & \cellcolor{blue!14}-7.0 & \cellcolor{blue!14}32.1 & \cellcolor{blue!14}5.1 & \cellcolor{blue!14}10.1 &  & 55.0 & \textbf{36.4} & 23.8 & \underline{70.0} &  & 30.8 \\
     & + SFT & $\times$ & \cellcolor{blue!14}12.1 & \cellcolor{blue!14}33.5 & \cellcolor{blue!14}9.9 & \cellcolor{blue!14}18.5 &  & 78.5 & 31.1 & 23.5 & 64.8 &  & 36.2 \\
    \rowcolor{gray!14}
      &  & $\checkmark$ & \cellcolor{blue!14}14.3 & \cellcolor{blue!14}34.5 & \cellcolor{blue!14}9.5 & \cellcolor{blue!14}19.4 &  & 78.7 & 32.5 & \underline{24.8} & \textbf{70.6} &  & 37.8 \\
     &    + DPO & $\times$ & \cellcolor{blue!14}27.9 & \cellcolor{blue!14}37.2 & \cellcolor{blue!14}13.3 & \cellcolor{blue!14}26.1 &  & 80.2 & 32.1 & 22.4 & 67.0 &  & 40.0 \\
    \rowcolor{gray!14}
      &  & $\checkmark$ & \cellcolor{blue!14}29.7 & \cellcolor{blue!14}45.1 & \cellcolor{blue!14}15.4 & \cellcolor{blue!14}30.1 &  & 81.6 & \underline{33.4} & \textbf{25.2} & \underline{70.5} &  & 43.0 \\
     &    + PPO   & $\times$ & \cellcolor{blue!14}\underline{43.4} & \cellcolor{blue!14}\underline{50.9} & \cellcolor{blue!14}\underline{17.8} & \cellcolor{blue!14}\underline{37.4} &  & \underline{83.2} & 32.4 & 21.1 & 69.5 &  & \underline{45.5} \\
    \rowcolor{gray!14}
      &  & $\checkmark$ & \cellcolor{blue!14}\textbf{46.8} & \cellcolor{blue!14}\textbf{58.2} & \cellcolor{blue!14}\textbf{23.0} & \cellcolor{blue!14}\textbf{42.7} &  & \textbf{85.3} & 33.2 & 24.1 & 68.7 &  & \textbf{48.5} \\
    \cmidrule{1-7} \cmidrule{9-12} \cmidrule{14-14}
    Qwen2.5-7B-Instruct &  & $\times$ & \cellcolor{blue!14}22.2 & \cellcolor{blue!14}37.1 & \cellcolor{blue!14}10.0 & \cellcolor{blue!14}23.1 &  & 49.8 & \underline{22.2} & \textbf{28.2} & \textbf{75.4} &  & 35.0 \\
     & + SFT & $\times$ & \cellcolor{blue!14}12.6 & \cellcolor{blue!14}28.6 & \cellcolor{blue!14}10.0 & \cellcolor{blue!14}17.1 &  & 65.4 & 21.1 & 19.0 & 67.2 &  & 32.0 \\
    \rowcolor{gray!14}
      &  & $\checkmark$ & \cellcolor{blue!14}18.7 & \cellcolor{blue!14}33.7 & \cellcolor{blue!14}10.9 & \cellcolor{blue!14}21.1 &  & 71.0 & 21.8 & 21.8 & 60.5 &  & 34.1 \\
     &   + DPO & $\times$ & \cellcolor{blue!14}21.1 & \cellcolor{blue!14}31.4 & \cellcolor{blue!14}11.4 & \cellcolor{blue!14}21.3 &  & 71.3 & 21.4 & 16.7 & 68.5 &  & 34.5 \\
    \rowcolor{gray!14}
      &  & $\checkmark$ & \cellcolor{blue!14}28.1 & \cellcolor{blue!14}37.6 & \cellcolor{blue!14}\underline{14.9} & \cellcolor{blue!14}\underline{26.9} &  & 73.7 & 22.0 & \underline{22.8} & 66.0 &  & 37.9 \\
     &    + PPO   & $\times$ & \cellcolor{blue!14}\underline{33.1} & \cellcolor{blue!14}\underline{37.4} & \cellcolor{blue!14}\underline{15.3} & \cellcolor{blue!14}\underline{28.6} &  & 71.1 & 22.1 & 19.0 & \underline{71.5} &  & \underline{38.5} \\
    \rowcolor{gray!14}
      &  & $\checkmark$ & \cellcolor{blue!14}\textbf{39.8} & \cellcolor{blue!14}\textbf{45.3} & \cellcolor{blue!14}\textbf{17.7} & \cellcolor{blue!14}\textbf{34.3} &  & \textbf{76.5} & \textbf{22.4} & 22.4 & 71.4 &  & \textbf{42.2} \\
    \midrule
    \multicolumn{14}{l}{\underline{\textit{Zero Training (No SFT)}}}\\
    Llama-3.1-8B & Base (nonthink tpl) & $\times$ & \cellcolor{blue!14}-87.6 & \cellcolor{blue!14}1.7 & \cellcolor{blue!14}0.8 & \cellcolor{blue!14}-28.4 &  & 30.2 & 25.2 & \underline{17.0} & 36.6 &  & 3.4 \\
    \rowcolor{gray!14}
      & Base (think tpl) & $\checkmark$ & \cellcolor{blue!14}-88.2 & \cellcolor{blue!14}1.6 & \cellcolor{blue!14}0.6 & \cellcolor{blue!14}-28.7 &  & 31.8 & 20.7 & 16.7 & 26.5 &  & 1.4 \\
      &  + DPO & $\times$ & \cellcolor{blue!14}-74.6 & \cellcolor{blue!14}2.4 & \cellcolor{blue!14}0.3 & \cellcolor{blue!14}-24.0 &  & 26.1 & \underline{32.5} & 13.3 & \textbf{47.3} &  & 6.8 \\
    \rowcolor{gray!14}
      &  & $\checkmark$ & \cellcolor{blue!14}-68.9 & \cellcolor{blue!14}4.5 & \cellcolor{blue!14}1.0 & \cellcolor{blue!14}-21.1 &  & 30.1 & \textbf{35.3} & 13.6 & 14.5 &  & 4.3 \\
      &   + PPO   & $\times$ & \cellcolor{blue!14}\underline{-1.2} & \cellcolor{blue!14}\underline{19.5} & \cellcolor{blue!14}\underline{3.5} & \cellcolor{blue!14}\underline{7.3} &  & \underline{52.9} & 28.0 & 15.3 & \underline{44.4} &  & \underline{23.2} \\
    \rowcolor{gray!14}
      &  & $\checkmark$ & \cellcolor{blue!14}\textbf{10.4} & \cellcolor{blue!14}\textbf{26.1} & \cellcolor{blue!14}\textbf{3.8} & \cellcolor{blue!14}\textbf{13.4} &  & \textbf{56.6} & 31.9 & \textbf{18.7} & 31.5 &  & \textbf{25.6} \\
    \cmidrule{1-7} \cmidrule{9-12} \cmidrule{14-14}
    Qwen2.5-7B & Base (nonthink tpl) & $\times$ & \cellcolor{blue!14}-65.8 & \cellcolor{blue!14}4.5 & \cellcolor{blue!14}2.0 & \cellcolor{blue!14}-19.8 &  & \underline{39.1} & 23.1 & \textbf{17.3} & 63.2 &  & 11.9 \\
    \rowcolor{gray!14}
      & Base (think tpl) & $\checkmark$ & \cellcolor{blue!14}-68.4 & \cellcolor{blue!14}4.4 & \cellcolor{blue!14}1.5 & \cellcolor{blue!14}-20.8 &  & 36.7 & 23.0 & \textbf{17.3} & 59.9 &  & 10.6 \\
      &   + DPO & $\times$ & \cellcolor{blue!14}-24.5 & \cellcolor{blue!14}21.7 & \cellcolor{blue!14}\textbf{5.4} & \cellcolor{blue!14}0.9 &  & 38.3 & \underline{25.0} & 16.0 & \textbf{69.3} &  & \underline{21.6} \\
    \rowcolor{gray!14}
      &  & $\checkmark$ & \cellcolor{blue!14}\textbf{-18.3} & \cellcolor{blue!14}\textbf{24.6} & \cellcolor{blue!14}\underline{4.9} & \cellcolor{blue!14}\textbf{3.7} &  & \textbf{40.4} & \textbf{25.3} & 14.6 & \underline{68.8} &  & \textbf{22.9} \\
      &  + PPO   & $\times$ & \cellcolor{blue!14}-70.0 & \cellcolor{blue!14}2.9 & \cellcolor{blue!14}1.2 & \cellcolor{blue!14}-22.0 &  & 36.9 & 21.9 & \underline{16.3} & 49.2 &  & 8.3 \\
    \rowcolor{gray!14}
      &  & $\checkmark$ & \cellcolor{blue!14}\underline{-65.8} & \cellcolor{blue!14}5.9 & \cellcolor{blue!14}\underline{1.8} & \cellcolor{blue!14}\underline{-19.4} &  & 35.1 & 22.9 & \underline{16.0} & 54.1 &  & 10.0 \\
    \bottomrule
    \vspace{-2.5em}
    \end{tabular}
    \endgroup
\end{table}

Comparing Tables~\ref{tab:grpo-benchmarks} and \ref{tab:dpo-ppo-benchmarks} certifies that GRPO usually weighs in at about 1--3 points better than PPO, and about 5pp better than DPO on average.
RLMT models retain (and indeed, extend) their edge over non-thinking baselines even with DPO and PPO.
Not only are they better at Chat and Creative Writing, but they also show relative gains on instruction following (IF\textsubscript{Ben}), factuality (PopQA) and world knowledge (MMLU\textsubscript{R}).
While preference optimization has often turned to DPO and PPO in recent years, we invite researchers to explore the trade-offs of using GRPO more in future work.

%% file: appendices/prompts.tex
\section{Prompts}
\label{app:prompts}

In this Appendix, we provide the prompts used for various aspects of the experiments.

\paragraph{Prompt used to sample the warm-start data.}
We used Gemini 2.5 Flash (\texttt{gemini-2.5-flash-preview-04-17}) with the following instruction appended to each user prompt:

\begin{center}
\fbox{\begin{minipage}{0.97\textwidth}\ttfamily
FORMAT: First showcase a detailed planning phase where you plan your response within <think>...</think> tags. Then produce the actual response within <response>...</response> tags. The content within the <think>...</think> tags should *not* refer to the fact that a planning phase was prompted - they should refer to the user prompt only.
\end{minipage}}
\end{center}

\paragraph{Model output formats and corresponding prompts.}
For the warm-started models, we used the following format for the thinking models' outputs:
\begin{center}
\fbox{\begin{minipage}{0.97\textwidth}\ttfamily
<think> Some thinking here </think>
Response here
\end{minipage}}
\end{center}
The non-thinking models converse as usual.

For the prompted base models, we used the following format to elicit a model response with thinking:
\begin{center}
\fbox{\begin{minipage}{0.97\textwidth}\ttfamily
A conversation between User and Assistant. Following the User's query, the Assistant first plans a response, and then provides the response. The internal reasoning process is enclosed within <think> </think> tags and the response is enclosed within <response> </response> tags, i.e., in the format <think> reasoning process here </think> <response> response here </response>.

User: <query> ...user text... </query>
Assistant:
\end{minipage}}
\end{center}
The corresponding prompt for the non-thinking models is:
\begin{center}
\fbox{\begin{minipage}{0.97\textwidth}\ttfamily
A conversation between User and Assistant. The user asks a question, and the assistant provides the user with a response. The response is enclosed within <response> </response> tags, i.e., <response> response here </response>.

User: <query> ...user text... </query>
Assistant:
\end{minipage}}
\end{center}

\paragraph{Prompts used for trait extraction.}
We use the following prompt for the extraction of the initial set of traits
\begin{center}
\fbox{\begin{minipage}{0.97\textwidth}\ttfamily
You are analyzing the hidden planning part produced by a model before its final answer. From the planning excerpt below, infer the key characteristics of how the planning is performed. Focus on the style and intent of the planning, not the specific content of the question. Return ONLY a compact JSON array (no extra text), where each element is a short string naming one characteristic. Aim for 3-8 distinct, non-redundant items.
\end{minipage}}
\end{center}

We then prompt the model to compare batches of traits from the two models as follows.
\begin{center}
\fbox{\begin{minipage}{0.97\textwidth}\ttfamily
You will compare planning styles for model A vs model B. You are given multiple examples. For each, you will see the user prompt and two lists: A\_plan and B\_plan. Identify 1-3 concise, consistent differences describing how A's planning differs from B's. Focus on stylistic/strategic patterns that recur across the provided examples. Return ONLY a JSON array of short difference statements (no extra text).
\end{minipage}}
\end{center}

To run the identified (and summarized) traits through each prompt and calculate win rates, we prompt GPT-4.1-mini as thus:
\begin{center}
\fbox{\begin{minipage}{0.97\textwidth}\ttfamily
You are given two hidden planning excerpts from two models: A and B. For each trait, decide which planning shows the trait MORE strongly: 'A', 'B', or 'tie'. Return ONLY a JSON object mapping trait\_keys to 'A', 'B', or 'tie' (lowercase also accepted). Output strictly a JSON object with these keys only, each value one of 'A', 'B', or 'tie'.
\end{minipage}}
\end{center}

%% file: appendices/star_po.tex
\section{Brief Overview of Preference Optimization Algorithms}
\label{app:brief-dpo-ppo-grpo}

For the benefit of readers, we provide a short overview of DPO, PPO, and GRPO here; interested readers should refer to the cited papers for more details.

\paragraph{Setup.}
Let $\pi_\theta$ be a policy LM with parameters $\theta$ that maps a prompt $x$ to a distribution over responses $y$.
We write $\pi_{\mathrm{ref}}$ for a frozen reference policy (often the SFT model), and $r(y,x)$ for a scalar reward coming from either
a reward model (\emph{preference-based}) or a verifier (\emph{verifiable/automatic}). We denote log-probabilities by $\log \pi(\cdot \mid \cdot)$ and use a KL penalty weight $\lambda$ and a clipping parameter $\varepsilon$ where applicable. 

\paragraph{DPO (Direct Preference Optimization).}~\citep{rafailov2023direct}
\emph{Objective.}
Given offline pairwise preference data $D \equiv \{(x, y^+, y^-)\}$ with a preferred $y^+$ and a dispreferred $y^-$, DPO optimizes a logistic preference objective that implicitly constrains the policy to stay close to $\pi_{\mathrm{ref}}$:
\begin{align*}
\mathcal{L}_{\mathrm{DPO}}(\theta)
= - \mathbb{E}_{(x,y^+,y^-) \sim \mathcal{D}}
\Big[
\log \sigma \big(
\beta \big[
(\log \pi_\theta(y^+\!\mid x) - \log \pi_\theta(y^-\!\mid x))\\
- (\log \pi_{\mathrm{ref}}(y^+\!\mid x) - \log \pi_{\mathrm{ref}}(y^-\!\mid x))
\big]
\big)
\Big],
\end{align*}
where $\sigma$ is the logistic function and $\beta$ is a temperature. 
 DPO is an \emph{offline} preference-optimization method: no on-policy rollout is required. 
 It reframes RLHF as a calibrated logit-difference classification against the reference, avoiding a learned value function and making training simple and stable.

\paragraph{PPO (Proximal Policy Optimization).}~\citep{schulman2017proximal}
\emph{Objective.}
At each iteration, sample responses $y \sim \pi_{\theta_{\mathrm{old}}}(\cdot\mid x)$, compute rewards $R = r(y,x)$ (often with a per-sample KL penalty to $\pi_{\mathrm{ref}}$), estimate token- or sequence-level advantages $A$, and optimize the clipped surrogate:
\begin{align*}
\mathcal{L}_{\mathrm{PPO}}(\theta)
= \mathbb{E}\Big[\min\big(
\rho_\theta A,\;
\mathrm{clip}(\rho_\theta, 1-\varepsilon, 1+\varepsilon) A
\big)\Big]
- \lambda \,\mathrm{KL}\!\left(\pi_\theta(\cdot\mid x)\,\|\,\pi_{\mathrm{ref}}(\cdot\mid x)\right)
\end{align*}
where
\begin{align*}
\rho_\theta=\frac{\pi_\theta(y\mid x)}{\pi_{\theta_{\mathrm{old}}}(y\mid x)}.
\end{align*}
PPO is an \emph{on-policy} algorithm: the model rolls out new samples, gets a reward (e.g., from a reward model), uses generalized advantage estimation (GAE,~\citet{schulman2016highdimensional}) to form $A$, and updates with a clipping rule for stability. KL-to-reference regularization maintains alignment and prevents reward hacking.

\paragraph{GRPO (Group Relative Policy Optimization).}~\citep{shao2024deepseekmath}
\emph{Objective.}
For each prompt $x$, GRPO generates a \emph{group} of $K$ candidates $\{y_i\}_{i=1}^K$ from $\pi_{\theta_{\mathrm{old}}}$, obtain rewards $r_i = r(y_i,x)$, and compute group-centered advantages
$
A_i \!=\! r_i - \frac{1}{K}\sum_{j=1}^K r_j
$ instead of employing GAE.
Update the policy with a PPO-style clipped objective using these $A_i$ (often without a learned critic), plus an optional KL penalty:
\begin{align}
\mathcal{L}_{\mathrm{GRPO}}(\theta)
= \mathbb{E}_{x}\Bigg[\frac{1}{K}\sum_{i=1}^K
\min\big(
\rho_\theta^{(i)} A_i,\;
\mathrm{clip}(\rho_\theta^{(i)}, 1-\varepsilon, 1+\varepsilon) A_i
\big)\Bigg]
- \lambda \,\mathrm{KL}\!\left(\pi_\theta(\cdot\mid x)\,\|\,\pi_{\mathrm{ref}}(\cdot\mid x)\right).
\end{align}
GRPO is \emph{on-policy} but avoids a learned value function by using a per-prompt baseline (the group mean reward). It is especially convenient when rewards are naturally comparable within a prompt (e.g., verifiable correctness or a shared reward model). In practice it pairs well with \emph{reasoning} rollouts (sample multiple candidates per prompt, keep relative scores, and update toward better-than-average ones).

%% file: appendices/faqs.tex
\section{Frequently Asked Questions}
\label{app:faqs}

This appendix discusses relevant questions and directions that might stem from RLMT.

\paragraph{Can RLMT and RLVR be combined to build models that excel at both verifiable and non-verifiable tasks?}

Appendix~\ref{sec:more_results} provides preliminary results by applying RLMT on top of OpenThinker2-7B, a model already strong on verifiable tasks via distillation from DeepSeek-R1. Adding RLMT improves performance on chat-oriented benchmarks, but the resulting model still falls short of standalone RLMT models on non-verifiable tasks.
These findings suggest that simply applying RLVR followed by RLMT in sequence is insufficient to obtain a model that excels across all domains. Training pipelines that combine or balance multiple reward sources more effectively---for instance, by mixing verifiable and non-verifiable prompts during RL training, or by using reward ensembles---represent an important direction for future work.

\paragraph{Why does long-CoT benefit chat and creative writing?}

Long-CoT benefits chat and creative writing because it allows the model to (1) plan narrative structure and ensure consistency with previous plot points, (2) refine presentation to avoid common pitfalls like bullet-point formatting or repetitive character descriptions, and (3) balance multiple requirements (word count, style, plot elements) before committing to a response. Figure~\ref{fig:qual_analysis} in the main text shows which specific traits were learned during RLMT training that contribute to higher creative writing scores, and Figure~\ref{fig:intro_exs} provides an example response demonstrating these improvements.

\paragraph{Why reward only the final response, instead of the thinking and response together?}

In RLMT, rewards are measured only on the actual response (the visible output after \texttt{</think>}), while the CoT traces are optimized implicitly through backpropagation. An alternative approach would be to reward both the thinking and response using a reward model. Table~\ref{tab:reward-thinking-response} compares these two approaches.

\begin{table}[h]
    \small
    \centering
    \caption{Comparison of rewarding only the response versus rewarding both thinking and response. Rewarding only the response leads to substantially better performance, particularly on chat benchmarks.}
    \label{tab:reward-thinking-response}
    \begingroup
    \renewcommand{\tabcolsep}{1.2mm}
    \begin{tabular}{lrrrrrrrrr}
        \toprule
        \textbf{\emph{Model}} & \textbf{\emph{WB}} & \textbf{\emph{AE2}} & \textbf{\emph{AH2}} & \textbf{\emph{CWv3}} & \textbf{\emph{PopQA}} & \textbf{\emph{IF\textsubscript{Ben}}} & \textbf{\emph{MMLU\textsubscript{R}}} & \textbf{\emph{Avg}} \\
        \midrule
        Llama-3.1-8B-RLMT & \textbf{38.1} & \textbf{52.3} & \textbf{15.9} & \textbf{80.9} & 30.3 & 15.6 & 61.7 & \textbf{42.1} \\
        \quad + reward thinking \& response & 19.8 & 36.9 & 9.2 & 77.8 & \textbf{30.7} & \textbf{18.0} & \textbf{62.1} & 36.4 \\
        \bottomrule
    \end{tabular}
    \endgroup
\end{table}

Rewarding both thinking and response leads to substantially worse performance, particularly on chat benchmarks (WB, AE2, AH2). This suggests that the reward model is not well-calibrated to evaluate internal reasoning traces, and that the model learns thinking patterns more effectively through the final output reward alone. 

\paragraph{Doesn't ``thinking'' bring with it excess cost for the additional tokens?}

Long CoT utilizes extra tokens, costing both time and compute. However, we believe that the trade-off is worth navigating, since (1) model providers are already deploying models that always think, such as OpenAI's o1, o3, o4, and o4-mini (2) Recent releases like GPT-5 (and sub-versions) route requests to different models based on complexity, which represents one approach to alleviating costs. Additionally, we observe that far shorter CoT traces are sufficient to achieve strong performance on chat benchmarks (as compared to math, where we often need thousands of tokens), suggesting that the trade-off is not as severe as one might think.

\paragraph{Why does performance on some benchmarks (MMLU, PopQA, MATH-500) not improve substantially?}

It is unlikely for RL to teach new knowledge to the model, rather than refine already existing behavior. 
For this reason, our primary goal for knowledge-intensive benchmarks like MMLU and PopQA is to avoid performance degradation rather than achieve large gains. 
For math benchmarks such as MATH-500 (Appendix~\ref{sub:additional_benchmarks}), we see that performance does increase significantly from the SFT model, but that RLMT achieves similar gains to non-thinking settings. We chalk this up to the fact that our reward model has been trained largely on general chat prompts, rather than math. RLMT relies on a reward model and hence inherits the biases of the methods that use reward models.
Nonetheless, note that prominent preference optimization methods like SimPO~\citep{meng2024simpo} show significant performance drops on math benchmarks, and hence achieving gains itself is noteworthy.